\documentclass{article} %
\usepackage{iclr2027_conference,times}

\usepackage{amsmath,amsfonts,bm}

\def\eqref#1{equation~\ref{#1}}

\def\1{\bm{1}}

\DeclareMathAlphabet{\mathsfit}{\encodingdefault}{\sfdefault}{m}{sl}
\SetMathAlphabet{\mathsfit}{bold}{\encodingdefault}{\sfdefault}{bx}{n}

\usepackage{hyperref}
\usepackage{url}
\usepackage{graphicx}
\usepackage{wrapfig}
\usepackage{booktabs}
\usepackage{colortbl}
\definecolor{loopedrow}{rgb}{0.92,0.96,1.0}
\usepackage{multirow}
\usepackage{microtype}
\usepackage{placeins}
\usepackage{soul}
\usepackage{subcaption}

\title{\modelname{}: Fully-Open Pre-trained Looped Transformers with Attention Residuals}

\author{Yuekun Yao$^{1,*}$ \quad Yupei Du$^{2}$ \quad Dawei Zhu$^{2}$ \quad Huan Sun$^{1,*}$ \\
$^{1}$The Ohio State University \quad $^{2}$Saarland University \\
$^{*}$Correspondence: \texttt{\{yao.1267,sun.397\}@osu.edu} \\
\noalign{\vskip 5pt}
{\small\hypersetup{hidelinks}\raisebox{-2pt}{\includegraphics[height=10pt]{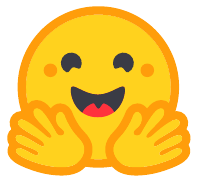}}\,\textsf{Models:}\quad
\href{https://huggingface.co/osunlp/ARLoop-1.3B}{\textcolor[HTML]{357985}{\texttt{ARLoop-1.2B}}} \quad
\href{https://huggingface.co/osunlp/ARLoop-0.6B}{\textcolor[HTML]{357985}{\texttt{ARLoop-0.6B}}}\quad
\href{https://huggingface.co/osunlp/ARLoop-baseline-0.6B}{\textcolor[HTML]{357985}{\texttt{PreNormLoop-0.6B}}}} \\
{\small\hypersetup{hidelinks}\raisebox{-2pt}{\includegraphics[height=10pt,trim=70 70 70 74,clip]{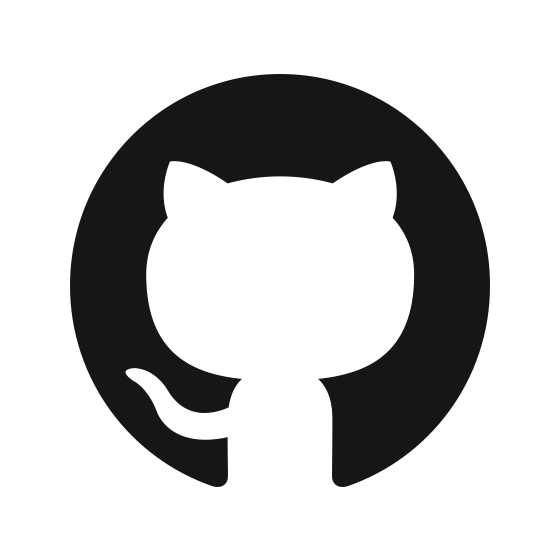}}\,\textsf{Code:}\quad
\href{https://github.com/OSU-NLP-Group/ARLoop}{\textcolor[HTML]{357985}{\texttt{github.com/OSU-NLP-Group/ARLoop}}}} \\
}

\newcommand{\modelname}{\textsc{ARLoop}}

\newcommand{\modelnamepl}{\textsc{ARLoop}\textsubscript{full}}
\newcommand{\modelnameplsw}{\textsc{ARLoop}\textsubscript{sw}}

\newcommand{\poissonSequenceModel}{\textsc{ARLOOP-Poisson-Seq}}
\newcommand{\poissonBatchModel}{\textsc{ARLOOP-Poisson-Batch}}

\newcommand{\prenorm}{\textsc{PreNormLoop}}
\newcommand{\prenormSmallModel}{\textsc{PreNormLoop-0.6B}}
\newcommand{\recurrentSmallModel}{\textsc{ARLoop-0.6B}}
\newcommand{\recurrentLargeModel}{\textsc{ARLoop-1.2B}}

\iclrfinalcopy %
\begin{document}

\maketitle

\begin{abstract}
Looped transformers increase effective depth by repeatedly applying a shared set of transformer layers, offering a parameter-efficient way to scale computation.
However, standard residual connections aggregate information across recurrent depth into a single stream, providing limited control over which intermediate representations from preceding layers should be reused.
In this paper, we introduce \modelname{}, a looped transformer architecture based on attention residuals \citep{kimi2026attentionresiduals}, where the input of each layer is selectively aggregated through attention over model depth.
Distinct from standard transformers, we find that looped transformers do not require access to the full residual history: 
sliding-window depth attention matches or outperforms full-history attention at larger recurrent depths.
Across controlled experiments with multiple parameter scales (140M and 370M) and recurrent depths ($R$$=$$1$ to $5$), \modelname{} consistently outperforms prior residual designs for looped transformers.
More importantly, we further scale \modelname{} to 0.6B and 1.2B parameters under a realistic LLM pretraining recipe, demonstrating that the gains transfer to realistic pretraining, with particularly strong improvements on math and coding tasks.
Our 1.2B model, pretrained on only 0.5T tokens with three recurrences, outperforms 
Huginn-3.5B (the only fully-open looped LLM as of our release date) 
across all evaluated tasks and is competitive with similar-sized non-looped 
models such as Qwen3.5-2B despite being trained on substantially less data.%
\footnote{We publicly release all artifacts, including training code, data processing code, intermediate checkpoints, and final \modelname\ models.}
\end{abstract}

\section{Introduction}
\label{sec:intro}
\begingroup
\setlength{\intextsep}{5pt}
\begin{figure}[!ht]
    \centering
    \includegraphics[width=\linewidth]{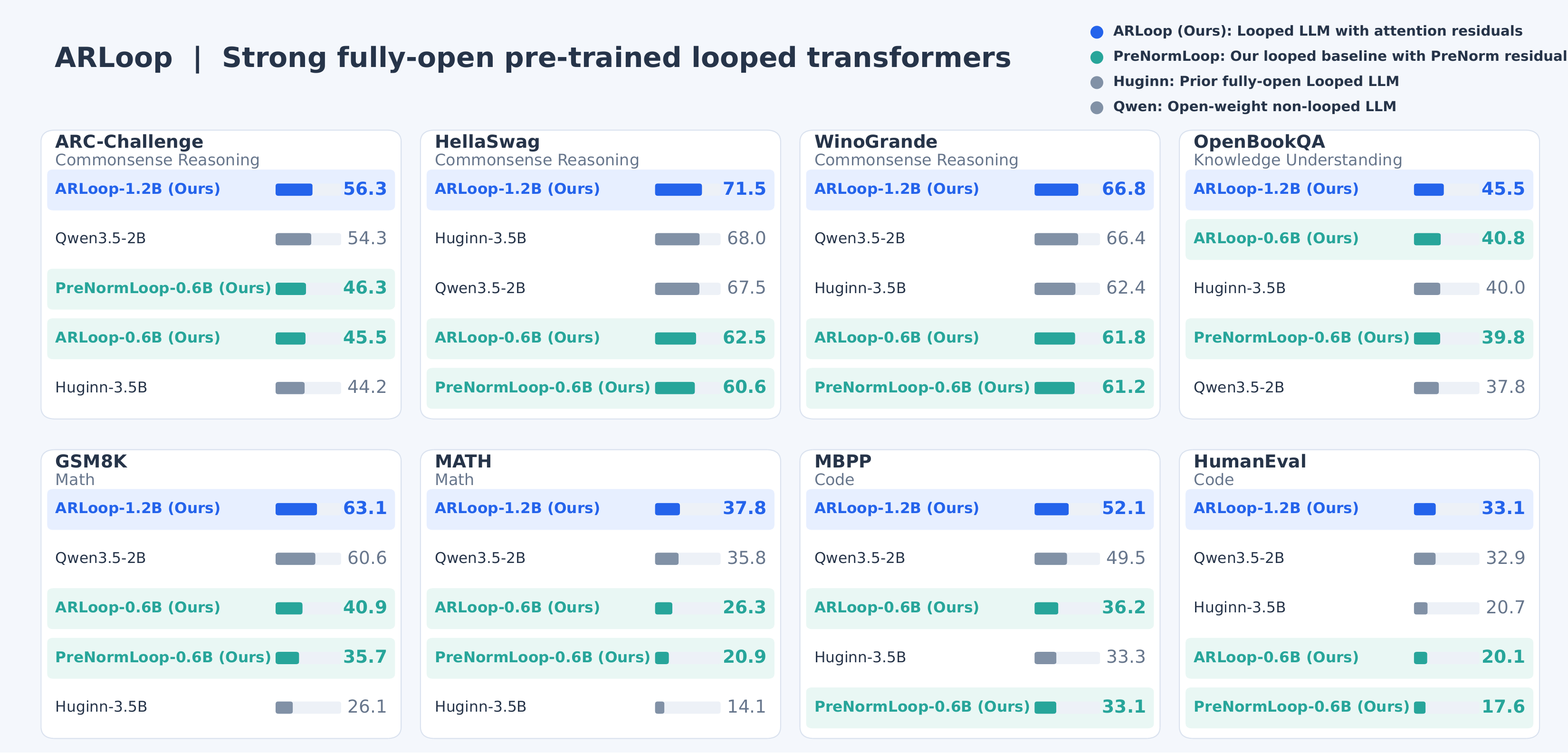}
    \captionsetup{font=footnotesize,skip=3pt}
    \caption{
    Performance of \modelname{} on general reasoning, knowledge, mathematics, and coding benchmarks. \recurrentLargeModel{} outperforms both Huginn-3.5B and Qwen3.5-2B on all eight benchmarks, while \recurrentSmallModel{} outperforms its parameter-matched \prenorm\ baseline on seven of eight benchmarks.
    }
    \label{fig:arloop_benchmark_overview}
\end{figure}
\endgroup

Looped transformers repeatedly apply a small set of shared transformer layers,
increasing effective depth without increasing the number of distinct parameters
\citep{dehghani2018universal,yang2024looped, saunshireasoning}.
This provides an appealing alternative to conventional depth scaling:
the same parameters can be reused for additional computation,
and the number of recurrences provides an explicit axis along which
training and inference compute can be varied. 
Recent work has demonstrated that such models can improve
generalization on controlled multi-hop reasoning tasks \citep{kohli2026loopthinkgeneralize}.
Meanwhile, growing interest has focused on scaling looped transformers
for parameter-efficient pretraining of large language models (LLMs)
\citep{geiping2025scaling,zhu2025scaling}.
A central question, however, is how to make looped transformers effectively
\emph{exploit the additional computation provided by recurrent depth}.

Increasing recurrent depth repeatedly produces new intermediate representations,
but standard transformer residual connections, such as PreNorm
\citep{xiong2020layernormalizationtransformerarchitecture},
provide only a limited mechanism for controlling how information produced
at earlier depths is reused.
Under PreNorm, residual updates from previous depths are accumulated
into a single residual stream with fixed unit coefficients.
As recurrence increases, more updates are accumulated into the same state,
while later computations have no explicit mechanism for selecting
which representations produced at earlier depths are useful.
Existing work has improved looped transformers through residual scaling
or normalization
\citep{geiping2025scaling,li2026deeploopdepthscalinglooped,
prairie2026parcaescalinglawsstable},
but limited systematic comparisons leave it unclear which residual design best enables models 
to make use of recurrent computation.
Moreover, these prior residual designs have mostly been evaluated under
relatively small-scale pretraining setups, leaving their effectiveness
on more complex generative tasks such as mathematical reasoning and
code generation underexplored.

Motivated by recent advances in attention residuals for standard transformers
\citep{kimi2026attentionresiduals},
we investigate how selective residual access should be adapted to recurrent depth.
Compared with PreNorm, attention residuals replace the uniform aggregation
of earlier residual updates with attention-weighted aggregation,
allowing each layer to selectively access representations from preceding depths.
Unlike standard transformers, however, recurrent depth repeatedly applies
the same parameters 
and can produce highly redundant representations
across recurrences \citep{zeitoun2026hyperlooptransformers, wang2026hrmtext}, raising the question of whether access to the full
residual history is necessary.

Our contributions are twofold.
First, we introduce \modelname{}, a looped transformer architecture
that adapts attention residuals to recurrent depth.
Across multiple parameter scales (140M and 370M) and recurrence depths
($R=1$ to $R=5$), \modelname{} consistently improves validation perplexity
over PreNorm and outperforms prior residual designs
\citep{geiping2025scaling,li2026deeploopdepthscalinglooped,
prairie2026parcaescalinglawsstable} under our rigorous controlled evaluation setting.
Distinct from standard transformers, we find that access to the full residual
history is unnecessary for looped transformers:
sliding-window depth attention matches or outperforms full-history attention
at larger recurrence depths.
Our analysis further shows that depth attention concentrates on recent
residual sources and that representations across recurrences are highly
redundant, explaining the effectiveness of sliding-window attention.

More importantly, we demonstrate that \modelname{} scales to work under a realistic LLM pretraining recipe.
Despite using a much smaller training budget (i.e.\ 0.5T tokens), our strongest \recurrentLargeModel{} substantially outperforms 
Huginn-3.5B \citep{geiping2025scaling} (the only fully-open looped LLM as of our release date)%
, and achieves performance competitive with strong non-looped LLMs such as Qwen3.5-2B.
At scale, we find that \modelname{} generally outperforms PreNorm, with especially promising improvements on math and coding tasks (e.g.\ +5.2 points on GSM8K and +3.1 points on MBPP).
We release training code and checkpoints, making \modelname{} fully-open for future study.

\section{Related work}

\paragraph{Looped transformers.}
Looped transformers, also known as recurrent-depth transformers or universal transformers \citep{dehghani2018universal, Lan2020ALBERT:, yang2024looped}, are widely studied as memory-efficient alternatives to transformers.
As they are increasingly scaled to LLM pretraining \citep{csordas2024moeut, geiping2025scaling, bae2025mixtureofrecursionslearningdynamicrecursive}, recent work has explored residual designs for better information propagation across recurrent depth, including Parcae, DeepLoop, and HyperLoop \citep{li2026deeploopdepthscalinglooped, zeitoun2026hyperlooptransformers, prairie2026parcaescalinglawsstable, knupp2026depthrecurrentattentionmixturesgiving}.
Compared with these approaches, \modelname{} consistently matches or outperforms them across parameter and recurrence budgets in our controlled experiments. We further show that these gains extend beyond language modeling perplexity to mathematical reasoning and code generation, which remain underexplored in prior work. 

\paragraph{Pretraining looped transformers at scale.}
Recent work has begun scaling looped transformers for pretraining.
\citet{geiping2025scaling} introduced Huginn, a 3.5B-parameter looped LLM trained with up to 64 recurrences, while \citet{zhu2025scaling} further scaled the training data with Ouro, a 1.4B model using up to four recurrences.
Subsequent work has explored looped LLMs with mixture-of-experts architectures such as Loopies \citep{gao2026looploopies}, SMELT \citep{wang2026smeltscalinglawscomputematched}, as well as larger data scales for advanced capabilities, including  Nanbeige4.2-3B \citep{lab2026nanbeige423bunlockingagenticcapabilities} and LoopCoder-v2 \citep{yang2026loopcoder}.
Despite these promising results, many looped LLMs release only model weights
or do not release weights at all, leaving the training and data recipe unclear.
The only exception is Huginn, which releases all training artifacts.
In particular, Huginn reports that PreNorm does not effectively exploit recurrent depth and instead relies on conservative optimization setups for stable training.
Our work systematically studies residual design for looped transformers and further validates the resulting architecture under a realistic LLM pretraining recipe, substantially outperforming Huginn despite using smaller parameter, data, and recurrence budgets.

\section{\modelname{}}
\label{sec:method}

We first describe the formulation of our looped transformer and then how we adapt attention residuals for it.
We refer to the resulting model as \modelname{}. 

\subsection{Preliminaries}\label{sec:looped_transformer}
\paragraph{Terminology.}
Throughout the paper, we use \textit{sublayer} to refer to either a self-attention or a feed-forward network (FFN) module,
\textit{transformer layer} to refer to a pair of self-attention and FFN sublayers,
and \textit{residual block} to refer to a group of consecutive sublayers used as a source in attention residuals, as introduced below.

\paragraph{Standard Transformer with PreNorm residual.}
We start by defining a standard decoder-only transformer \citep{NIPS2017_3f5ee243, radford2019language} with a PreNorm residual stream.
Let $x$ be a sequence of tokens, and let
$\bm{h}_{0} \in \mathbb{R}^{n \times d}$ denote their embeddings.
A standard transformer consists of $L$ layers, each of which contains a self-attention sublayer followed by an FFN.
For the $i$-th sublayer, its input $h_{i-1}$ and output $h_i$ can be written as
\begin{align}
{\bm{h}}_{i}
&= \bm{h}_{i-1}
+ {f_{i}}(\operatorname{Norm}(\bm{h}_{i-1})), \quad i = 1,\ldots,2L
\label{eq:sublayer}
\end{align}
where $f_i$ denotes either a self-attention or an FFN function.
We can then define $\bm{h_i} = \mathcal{F}_i(h_{i-1})$, and the entire transformer can be written as
$
\bm{y}
=
\operatorname{LMHead}
\circ \mathcal{F}_{2L}
\circ \cdots
\circ \mathcal{F}_1
(\bm{h}_{0}),
$
where $\operatorname{LMHead}$ denotes the language-modeling head that maps the final hidden states to output logits.

\paragraph{Looped transformer.}
Following the notation above, an $L$-layer looped transformer reuses the same transformer stack $\mathcal{S}$ (e.g.\ $\mathcal{F}_4 \circ \mathcal{F}_3 \circ \mathcal{F}_2 \circ \mathcal{F}_1$) for $R$ recurrences.
Although $\mathcal{S}$ can consist of any subset of the $L$ distinct transformer blocks, in this paper we only consider looping the full stack of transformer blocks (i.e.\ $\mathcal{S}=\mathcal{F}_{2L} \circ \cdots \circ \mathcal{F}_1$).
Hence, our looped transformer can be written as
$
\bm{y}
=
\operatorname{LMHead}
\circ \mathcal{S}^{(R)}
\circ \cdots
\circ \mathcal{S}^{(1)}
(\bm{h}_{0}),
\label{eq}
$
where $\mathcal{S}^{(1)}=\cdots=\mathcal{S}^{(R)}=\mathcal{S}$,
since the same transformer stack is shared across all recurrences.
The resulting unrolled network therefore has an effective depth of $2LR$ sublayers,
while containing only $2L$ distinct transformer sublayers.

\paragraph{Attention Residuals.}
Unrolling Equation \ref{eq:sublayer} gives us 
$
\bm{h}_i
= \bm{h}_0 + \sum_{j=1}^{i}\bm{f}_j(\operatorname{Norm}(\bm{h}_{j-1}))
\label{eq}
$, which implies that under PreNorm, every residual update produced at an earlier sublayer contributes to all subsequent residual states with a fixed unit coefficient.
Attention Residuals \citep{kimi2026attentionresiduals} replace this fixed aggregation with attention-based aggregation over residual sources across depth.
Specifically, we define the input embedding as $\bm{z}_0 := \bm{h}_0$,
and formulate attention residuals as
\begin{align}
\bm{z}_j
&= f_j\left(\operatorname{Norm}(\bm{h}_{j-1})\right),
\qquad
\alpha_{i,j}
= \operatorname{Softmax}_{j\in\{0,\ldots,i\}}\left(
\bm{q}_i^\top \operatorname{Norm}(\bm{z}_j)
\right), \label{eq:res_update} \\
\bm{h}_i
&= \alpha_{i,0}\bm{h}_0
+ \sum_{j=1}^{i}\alpha_{i,j}\bm{z}_j.
\label{eq:attnres}
\end{align}
where $\bm{q}_i$ is a learned query vector of the $i$-th sublayer.
Compared with PreNorm (i.e.\ Equation~\ref{eq:sublayer}), Attention Residuals (i.e.\ Equation~\ref{eq:attnres}) assign an attention-based coefficient $\alpha_{i,j}$ to each residual update $\bm{z}_j$, rather than aggregating all residual updates with unit coefficients.

Equation~\ref{eq:attnres} introduces a new residual source after every
sublayer, causing the number of sources to grow linearly with effective
depth. Following \citet{kimi2026attentionresiduals}, we therefore use
\emph{block attention residuals}, which group consecutive sublayers and
aggregate their residual updates into block-level representations.
Specifically, for a block of $S$ consecutive sublayers, we define its
representation as
\begin{equation}
    \bm{b}_k
    =
    \sum_{j=(k-1)S+1}^{kS} \bm{z}_j 
    \label{eq:block}
\end{equation}
At each sublayer, attention is computed over the input embedding,
the representations of preceding completed blocks, and the partial sum
of residual updates within the current block.
Thus, residual updates are accumulated additively within a block as in
PreNorm, while attention-based aggregation is performed across blocks.
This reduces the number of stored residual sources from individual
sublayer outputs to block-level representations, substantially reducing
the 
overhead of full attention residuals.
We use block attention residuals for all \modelname{} 
models unless stated
otherwise.

\begin{figure*}[t]
    \centering
    \includegraphics[width=0.98\textwidth]{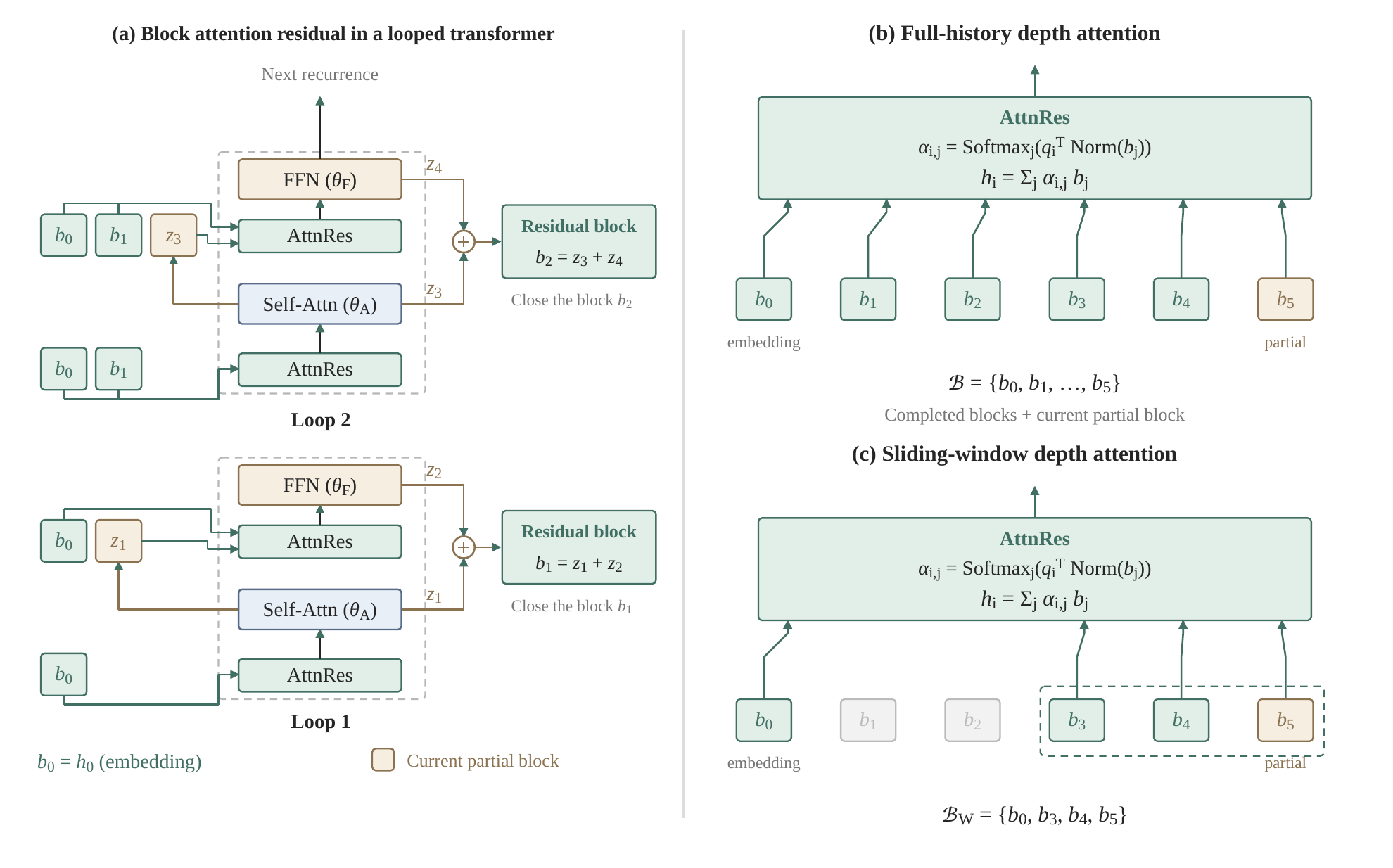}
    \caption{An example illustrating attention residuals in looped transformers.
\textbf{(a)} Each loop contains one transformer layer (i.e.\ a self-attention and an FFN), with parameters shared across loops.
We group every $S=2$ sublayers into a residual block, yielding
$b_1=z_1+z_2$ and $b_2=z_3+z_4$.
\textbf{(b)} Full-history depth attention \citep{kimi2026attentionresiduals} accesses all available residual sources.
\textbf{(c)} Sliding-window depth attention retains the embedding and a window of recent residual sources. 
}
    \label{fig:attnres_variants}
\end{figure*}

\subsection{Applying Attention Residuals to Looped Transformer}
\label{sec:attnres} 
Since attention residuals only change how each sublayer aggregates representations
from preceding depths, they can be naturally applied to looped transformers.
Figure~\ref{fig:attnres_variants}a illustrates a simplified example in which
each recurrence contains two sublayers.
As the depth grows, residual updates are grouped into one or more residual
blocks following Equation \ref{eq:block}.
Each effective sublayer has its own learned query vector
(i.e.\ $q_i$ in Figure~\ref{fig:attnres_variants}b),
which attends over the residual sources available at that depth.
Compared with PreNorm-based looped transformers, this allows sublayers at
deeper recurrences to selectively access representations produced at earlier
recurrent depths.

\paragraph{Sliding window depth attention.}

Standard attention residuals allow each depth-attention operation to access
all preceding residual blocks as candidate sources, which we refer to as
\emph{full-history depth attention}.
In a looped transformer, however, the same set of parameters is
repeatedly applied across recurrences, potentially introducing redundancy among
representations produced at different recurrent depths.
Indeed, prior work has observed high similarity between hidden states across
recurrences in looped transformers
\citep{zeitoun2026hyperlooptransformers}.
This motivates us to ask whether access to the full residual history is necessary,
or whether recent residual states already provide sufficient information.

Specifically, we study if sliding-window depth attention (Figure \ref{fig:attnres_variants}c) can match full-history depth attention.
Let
$\mathcal{B}=\{b_0,b_1,\ldots,b_m\}$
denote the residual sources available to a depth-attention operation,
where $b_0$ denotes the input embedding and $b_m$ is the most recent residual block.
Given a window size $W$, sliding-window depth attention restricts the source set to
$
    \mathcal{B}_{W}
    =
    \{b_0, b_{\max(1,m-W+1)},\ldots,b_m\}.
$
Thus, the input embedding remains directly accessible at every depth,
while depth attention is performed only over the $W$ most recent residual blocks.
Note that sliding-window depth attention has also been studied in standard transformers,
where it reduces the gains of attention residuals over PreNorm
\citep{kimi2026attentionresiduals}.
In contrast, as we will show in Section~\ref{sec:exp1},
sliding-window depth attention can match and even outperform full-history
attention in looped transformers, particularly at larger recurrence depths.
We therefore make sliding-window depth attention the default in \modelname{} for $R>2$.

\section{Controlled Evaluation of \modelname{}}
\label{sec:exp1}

We first use controlled experiments to evaluate how effectively  \modelname{} reuses information across recurrent depth, examining both its benefits for language modeling and the contribution of its key design choices.

\paragraph{General setup.}
We evaluate \modelname{} under a controlled setup using 140M- and 370M-parameter
decoder-only transformers with 6 and 12 transformer layers, respectively.
Following \citet{prairie2026parcaescalinglawsstable}, we pretrain these models
on 11.2B and 29.6B tokens from FineWeb-Edu \citep{penedo2024fineweb},
respectively, with a context length of 2048.
Following the general architecture of \citet{nanochat}, each transformer layer uses
multi-head attention and a standard MLP.
For optimization, we use AdamW \citep{kingma2017adammethodstochasticoptimization} with a cosine learning rate schedule.
Complete architecture and training details are provided in Appendix~\ref{app:exp1_setup}.

\paragraph{\modelname{} training setup.}
For all \modelname{} models, we use fixed recurrence, such that
the number of recurrences $R$ is fixed throughout training.
For the 140M and 370M models, we group every 6 and 12 consecutive sublayers,
respectively, into a residual block.
Since each recurrence contains 12 sublayers at 140M and 24 sublayers at 370M,
this yields two residual blocks per recurrence at both scales.
Following the design in Section~\ref{sec:attnres}, we use full-history depth
attention for $R\leq2$ and sliding-window depth attention for $R>2$.
For the sliding-window variant, we retain the input embedding and the three
most recent residual blocks.

\paragraph{Baselines.}
We use a PreNorm-based looped transformer with the residual in Equation~\ref{eq:sublayer} as our default baseline,
which we call \prenorm.
We also compare against previously proposed residual designs
for looped transformers, including Sandwich Norm
\citep{geiping2025scaling,zhu2025scaling},
DeepLoop \citep{li2026deeploopdepthscalinglooped},
and Parcae \citep{prairie2026parcaescalinglawsstable}, both with and without input injection \citep{bansal2022end}.
We provide the formulations of these baselines in
Appendix~\ref{app:baselines}.
We train all models with a fixed number of recurrences $R=1,3,5$ at the 140M scale and $R=1,3$ at the 370M scale due to computational cost.
We implement and train these methods under a unified training setup,
keeping all training hyperparameters fixed across methods
except for the learning rate.
Details of learning-rate tuning are provided in
Appendix~\ref{app:exp1_setup}.

Besides comparison with prior methods, we also train two variants of \modelname{} to better understand our method.
Specifically, we train (1) \modelnamepl{}, which applies full-history depth attention; and (2) \modelnameplsw{}, which applies sliding-window depth attention, and compare them with the \prenorm\ baseline.
We train each model with a fixed number of recurrences, using one model for each
$R \in \{2,3,4\}$.

\paragraph{Evaluation.}
We use perplexity (PPL) on a held-out validation set of 10M tokens from Fineweb-Edu as the main evaluation metric.
In addition, we also evaluate performance on a broad set of downstream tasks.
Our \emph{downstream} aggregate is the unweighted mean over ARC-Challenge, ARC-Easy \citep{clark2018thinksolvedquestionanswering},
HellaSwag \citep{zellers2019hellaswagmachinereallyfinish}, LAMBADA under both the OpenAI and standard formulations \citep{paperno-etal-2016-lambada}, OpenBookQA \citep{mihaylov2018suitarmorconductelectricity}, PIQA \citep{bisk2019piqareasoningphysicalcommonsense}, RACE \citep{lai-etal-2017-race}, SciQ \citep{welbl-etal-2017-crowdsourcing}, and WinoGrande \citep{sakaguchi2019winograndeadversarialwinogradschema}.
We use normalized accuracy for HellaSwag, OpenBookQA, and PIQA, and accuracy for the remaining tasks.
Complete evaluation details are provided in Appendix~\ref{app:controlled_evaluation}.

\subsection{Results} \label{sec:main_small_results}

\begin{table}[t]
\centering
\small
\setlength{\tabcolsep}{5pt}
\resizebox{0.85\linewidth}{!}{%
\begin{tabular}{@{}lrrr|rr@{}}
\toprule
Model & \multicolumn{3}{c|}{140M} & \multicolumn{2}{c}{370M} \\
\cmidrule(lr){2-4}\cmidrule(lr){5-6}
& $R=1$ & $R=3$ & $R=5$ & $R=1$ & $R=3$ \\
\midrule
Sandwich-Norm \citep{geiping2025scaling} & 18.98 & 17.19 & 16.59 & 13.92 & 12.97 \\
DeepLoop \citep{li2026deeploopdepthscalinglooped} & 18.72 & 17.05 & 16.52 & 13.69 & 12.78 \\
Parcae w/o input injection \citep{prairie2026parcaescalinglawsstable} & 18.37 & 16.60 & 16.01 & 13.55 & 12.31 \\
Parcae w. input injection \citep{prairie2026parcaescalinglawsstable} & 18.37 & 16.55 & 15.96 & 13.55 & 12.30 \\
\prenorm\ & 18.37 & 16.91 & 16.34 & 13.55 & 12.48 \\
\modelname{} \textsc{} & \textbf{18.26} & \textbf{16.38} & \textbf{15.84} & \textbf{13.32} & \textbf{12.17} \\
\bottomrule
\end{tabular}
}
\caption{Validation perplexity ($\downarrow$) for models at the 140M and 370M scales. 
}
\label{tab:related-model-val-ppl}
\end{table}

\paragraph{\modelname{} consistently outperforms existing residual designs in language modeling.}
We first compare the validation perplexity of \modelname{} in Table~\ref{tab:related-model-val-ppl} against residual designs proposed in prior work.
Across different parameter, data, and recurrence budgets, \modelname{} consistently achieves the lowest validation perplexity among \prenorm\ and the other residual designs, demonstrating the effectiveness of selective access to representations from earlier depths.
Compared with \prenorm, \modelname{} yields larger gains at deeper recurrences (e.g.\ $-0.11$ PPL at $R=1$ versus $-0.53$ at $R=3$), 
suggesting that \modelname{} makes more effective use of the additional computation provided by recurrent depth.

\begin{wraptable}{r}{0.6\textwidth}
\centering
\scriptsize
\setlength{\tabcolsep}{2.5pt}
\renewcommand{\arraystretch}{1.12}
\resizebox{\linewidth}{!}{%
\begin{tabular}{@{}ccrrr|rrr@{}}
\toprule
\multirow{2}{*}{Scale}
& \multirow{2}{*}{$R$}
& \multicolumn{3}{c|}{PPL $\downarrow$}
& \multicolumn{3}{c}{Downstream $\uparrow$} \\
\cmidrule(lr){3-5}\cmidrule(lr){6-8}
& & \prenorm & \modelnamepl{} & \modelnameplsw{}
  & \prenorm & \modelnamepl{} & \modelnameplsw{} \\
\midrule
\multirow{3}{*}{140M}
& 2 & 17.48 & \textbf{17.27} & 17.30
    & 42.22 & 41.96 & \textbf{42.27} \\
& 3 & 16.91 & 16.46 & \textbf{16.38}
    & 42.72 & \textbf{43.69} & 43.09 \\
& 4 & 16.59 & 16.07 & \textbf{16.03}
    & 43.27 & \textbf{44.09} & 43.73 \\
\midrule
\multirow{3}{*}{370M}
& 2 & 12.83 & {12.68} & \textbf{12.55}
    & 48.98 & 48.70 & \textbf{49.25} \\
& 3 & 12.48 & 12.27 & \textbf{12.17}
    & 49.29 & 50.23 & \textbf{50.67} \\
& 4 & 12.27 & 12.01 & \textbf{11.99}
    & 50.19 & \textbf{50.93} & 50.76 \\
\bottomrule
\end{tabular}%
}
\caption{Validation perplexity and average zero-shot downstream
accuracy (\%).
Bold marks the best result for each metric within each row.}
\label{tab:ablation}
\end{wraptable}

\paragraph{Sliding-window attention matches full-history attention at shallow recurrence, and even outperforms it in perplexity at deeper recurrences.}
We next ask whether looped transformers need direct access to the full residual
history as recurrent depth increases.
Table~\ref{tab:ablation} compares PreNorm (\prenorm), full-history depth attention
(\modelnamepl{}), and sliding-window depth attention (\modelnameplsw{}).
In terms of perplexity, although \modelnameplsw{} performs worse than full-history depth attention
(i.e.\ \modelnamepl{}) at $R=2$ of 140M scale, it achieves the best performance at larger recurrence depths and large parameter scales.
Both variants also generally outperform \prenorm\ in downstream accuracy, suggesting that the effectiveness of our method extends to general knowledge and reasoning tasks.
\textit{Given that the sliding-window variant also reduces computational overhead, we therefore adopt
the sliding-window variant \modelnameplsw{}
as the default \modelname{}}.

\begin{figure}[t]
    \centering
    \begin{subfigure}[t]{0.5\linewidth}
        \centering
        \includegraphics[height=0.28\textheight]{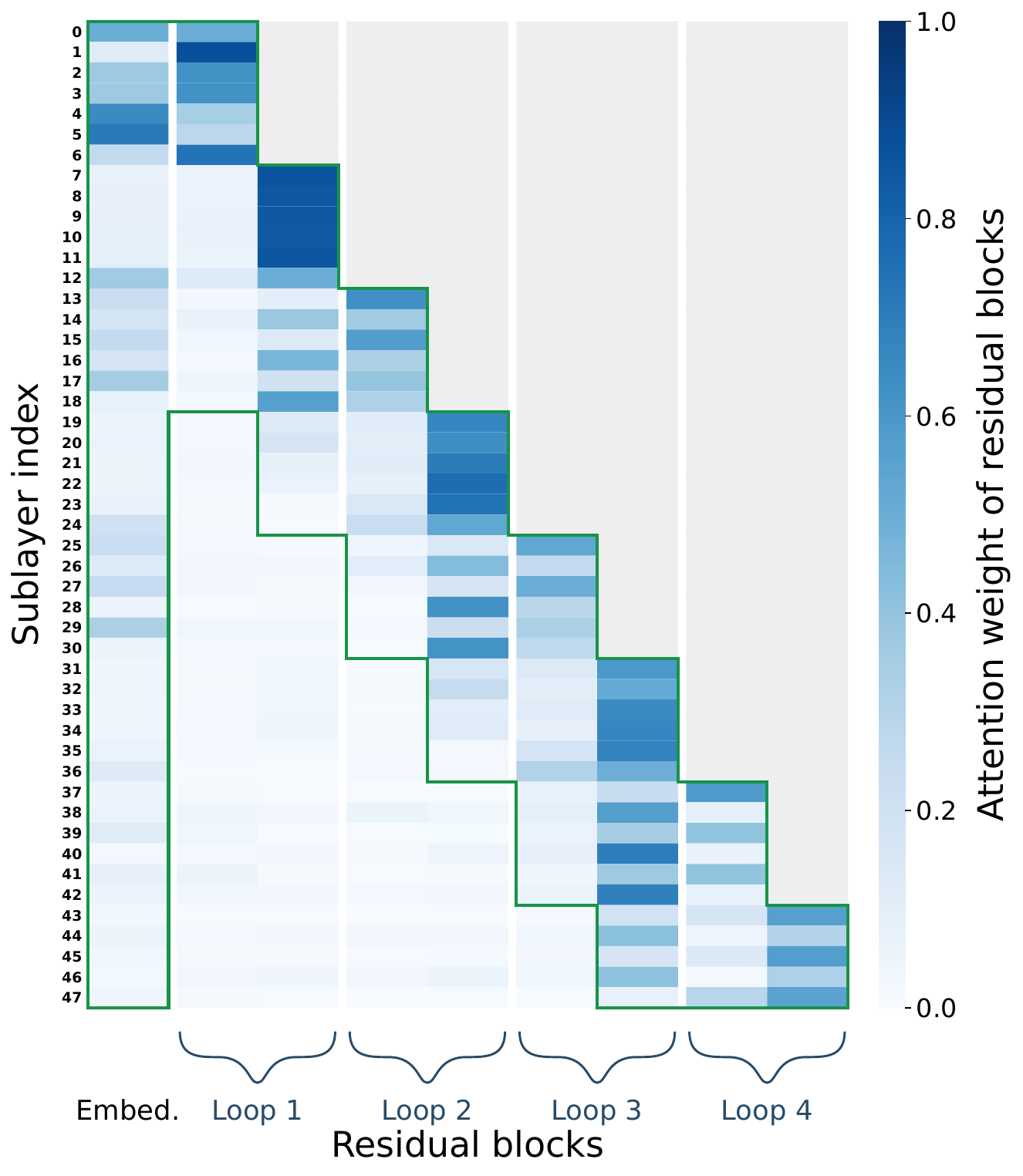}
        \label{fig:attnres_l4_sources}
    \end{subfigure}
    \hfill
    \begin{subfigure}[t]{0.48\linewidth}
        \centering
        \includegraphics[height=0.28\textheight]{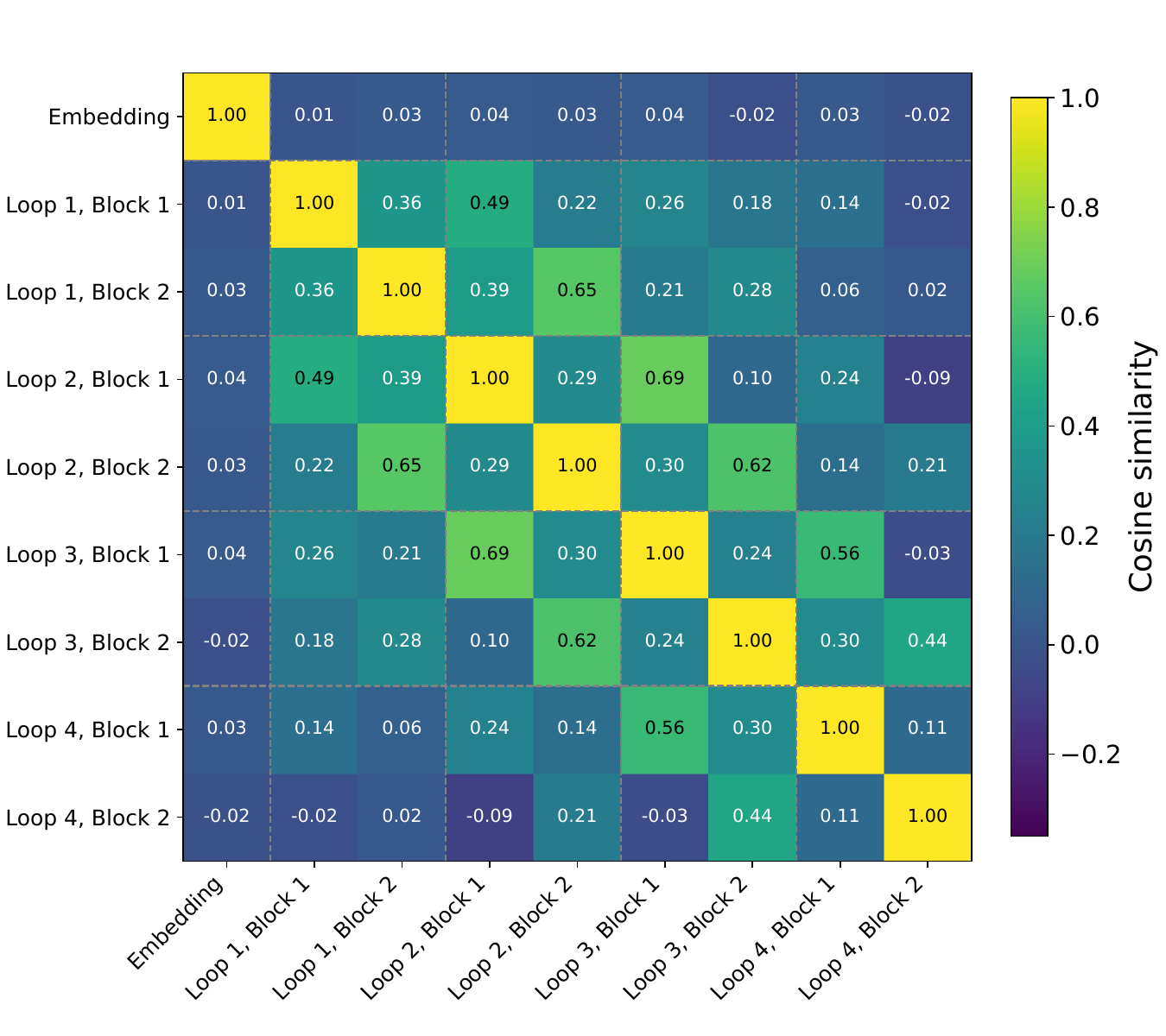}
        \label{fig:attnres_l4_blocks}
    \end{subfigure}
\caption{
\textbf{Left}: Attention weights over residual sources (i.e.\ blocks) at each sublayer for the 140M \modelnamepl{} model with $R=4$, averaged over the validation set.
Green lines indicate the sliding windows used in the corresponding \modelnameplsw{} variant.
\textbf{Right}: Pairwise cosine similarity between residual blocks for the same model. 
}
    \label{fig:attnres_140m_r4_sources}
\end{figure}

\paragraph{Why sliding-window depth attention works for looped transformers.}
To understand why recent residual blocks are sufficient, we visualize the
attention weights assigned to each residual source by \modelnamepl{} with
$R=4$ in Figure~\ref{fig:attnres_140m_r4_sources} (Left).
Depth attention places most of its mass on recent residual sources 
(e.g., at sublayer 47, the three most recent residual blocks
and the embedding block receive 96\% of the total attention weight),
suggesting that direct access to distant residual history contributes relatively
little to the representations used at deeper recurrences.
We hypothesize that this locality arises from parameter sharing across recurrences: repeatedly applying the same transformer layers produces redundant
representations, while more recent sources have already undergone refinement through additional
recurrent computation.
Consequently, retaining direct access to distant residual sources may provide little additional information and could even introduce unnecessary interference.

To examine this hypothesis, we further visualize the pairwise cosine similarity between residual blocks in Figure~\ref{fig:attnres_140m_r4_sources} (Right).
Residual sources across recurrences exhibit high similarity, supporting the presence of substantial redundancy across recurrent depth.
We observe the same pattern across other parameter scales and recurrence settings, with complete results provided in Appendix~\ref{app:sliding_window_analysis}.
Importantly, this behavior differs from a standard transformer with matched
effective depth, where representations exhibit substantially lower cross-depth similarity, and
applying the same sliding-window restriction degrades performance
(Appendix~\ref{app:matched_depth}).
Together, these results suggest that local depth attention is particularly
well suited to the recurrent-depth structure of looped transformers.

\section{Scaling to Practical  Pretraining of Looped Transformers}
\label{sec:scaling_results}

Having established the effectiveness of \modelname{} in controlled experiments, we now scale it to billion-parameter models, bringing together carefully curated pretraining data mixtures and a multi-stage training strategy to build strong, fully-open \modelname{}-based LLMs.

\paragraph{Setup.}
We adopt a pretraining recipe similar to SmolLM
\citep{allal2025smollm2,huggingfacetb2025smollm3}.
Compared with the controlled experiments in Section~\ref{sec:exp1},
we replace multi-head attention with grouped-query attention \citep{ainslie2023gqatraininggeneralizedmultiquery}
and use SwiGLU \citep{shazeer2020gluvariantsimprovetransformer} in place of the standard FFN sublayers to better align with modern LLM architectures \citep{qwen2025qwen3}. 
For pretraining data, we use FineWeb-Edu \citep{penedo2024fineweb}, FineMath, and Stack-Edu \citep{allal2025smollm2} as the primary sources.
We use the same cosine learning rate schedule and adopt a multi-stage training
strategy, introducing higher-quality data in later stages following
\citet{huggingfacetb2025smollm3} and \citet{allenai2025olmo3base}.
We also extend the context length to 4096 tokens.
Such a realistic pretraining recipe enables us to evaluate our models not only on general knowledge and reasoning tasks, but also on mathematical reasoning and code generation.
We provide complete architecture, training, and data-mixture details in
Appendix~\ref{app:data_mixture}.

Using this recipe, we train three models:
(1) \prenormSmallModel{}, a 0.6B \prenorm\ model with 18 transformer layers and $R=3$, trained on 180B tokens;
(2) \recurrentSmallModel{}, a 0.6B \modelname{} model with the same architecture, recurrence, and training budget; and
(3) \recurrentLargeModel{}, a 1.2B \modelname{} model with 24 transformer layers and $R=3$, trained on 500B tokens.
For both \modelname{} models, we adopt the sliding-window configuration from Section~\ref{sec:exp1}.

\paragraph{Evaluation.}
We use ARC-Challenge \citep{clark2018thinksolvedquestionanswering}, HellaSwag \citep{zellers2019hellaswagmachinereallyfinish}, WinoGrande \citep{sakaguchi2019winograndeadversarialwinogradschema}, and OpenBookQA \citep{mihaylov2018suitarmorconductelectricity}
to evaluate general knowledge and reasoning;
GSM8K \citep{cobbe2021training} and MATH \citep{hendrycks2021math, lewkowycz2022solvingquantitativereasoningproblems}
for mathematical reasoning;
and MBPP \citep{austin2021program} and HumanEval \citep{chen2021evaluating} for code generation.
Complete evaluation details are provided in Appendix~\ref{app:large_scale_evaluation}.

\paragraph{Baselines.}
We additionally compare against both existing looped language models
and strong non-looped LLMs of similar or larger sizes.
The recurrent-depth baselines include Huginn \citep{geiping2025scaling}
and Ouro \citep{zhu2025scaling},
while the non-looped open models include Qwen3.5-2B \citep{qwen2026qwen35base},
Gemma~3-4B \citep{gemmateam2025gemma3},
OLMo~2-7B and OLMo~3-7B \citep{olmo2024furious,allenai2025olmo3base}.
The non-looped open models are included mainly as references rather than for head-to-head comparison.
For all of these models, we re-evaluate the corresponding base checkpoints
using the same evaluation harness to ensure consistent evaluation.
Among these models, 
we particularly focus on Huginn because it is a fully-open looped LLM trained with a comparable data scale, but with substantially larger parameter and recurrence budgets (e.g.\ 3.5B parameters and up to 64 recurrences).

Note that we do not compare with other residual methods such as Parcae or DeepLoop \citep{prairie2026parcaescalinglawsstable, li2026deeploopdepthscalinglooped, zeitoun2026hyperlooptransformers} in this realistic pretraining comparison.
Such previous work generally validates their methods only at small data scales (e.g.\ below 50B tokens) or with a single pretraining source (e.g.\ FineWeb-Edu), leaving the capabilities of the resulting models on more complex tasks such as math and coding open.
Our contribution specifically targets this question, aiming to show \modelname{} remains effective on such generative tasks under realistic pretraining settings.
\begin{table}[t]
    \centering
    \small
    \setlength{\tabcolsep}{2pt}
    \resizebox{\linewidth}{!}{%
    \begin{tabular}{lcrrrrrrrrrr}
        \toprule
        Model & Looped & {FLOPs} & Data (T) & ARC-C & Hella. & Wino. & OBQA & GSM8K & MATH & MBPP & HumanEval \\
        \midrule
        \multicolumn{12}{c}{\textbf{Open-weight LLMs}} \\
        Qwen3.5-2B 
        & None & 432 & 36 & 54.3 & 67.5 & 66.4 & 37.8 & 60.6 & 35.8 & 49.5 & 32.9 \\
        Gemma~3-4B 
        & None & 96 & 4.0 & 58.4 & 77.7 & 72.3 & 43.0 & 38.1 & 25.4 & 60.1 & 34.2 \\
         \rowcolor{loopedrow}
         Ouro-1.4B 
         & 4 & 259 & 7.7 & 59.9 & 74.4 & 70.5 & 41.4 & 78.2 & 70.5 & 68.3 & 52.4 \\
        
        \midrule
        \multicolumn{12}{c}{\textbf{Fully-open LLMs}} \\
        OLMo~2-7B 
        & None & 180 & 4.1 & 64.6 & 82.0 & 76.8 & 45.6 & 68.1 & 20.9 & 32.0 & 12.2 \\
        OLMo~3-7B 
        & None & 252 & 5.9 & 61.0 & 76.4 & 73.8 & 42.4 & 73.8 & 43.6 & 66.1 & 14.0 \\
        \rowcolor{loopedrow}
        Huginn-3.5B 
        & 32 & 28 & 0.18 & 33.6 & 50.6 & 52.7 & 33.4 & 5.1 & 3.7 & 14.8 & 10.3 \\
        \rowcolor{loopedrow}
        Huginn-3.5B 
        & 32 & 127 & 0.8 & 44.2 & 68.0 & 62.4 & 40.0 & 26.1 & 14.1 & 33.3 & 20.7 \\
        \rowcolor{loopedrow}
        \prenormSmallModel{} (Ours) & 3 & 2 & 0.18 & 46.3 & 60.6 & 61.2 & 39.8 & 35.7 & 20.9 & 33.1 & 17.6 \\
        \rowcolor{loopedrow}
        \recurrentSmallModel{} (Ours) & 3 & 2 & 0.18 & 45.5 & 62.5 & 61.8 & 40.8 & 40.9 & 26.3 & 36.2 & 20.1 \\
        \rowcolor{loopedrow}
        \recurrentLargeModel{} (Ours) & 3 & 10 & 0.5 & 56.3 & 71.5 & 66.8 & 45.5 & 63.1 & 37.8 & 52.1 & 33.1 \\

        \bottomrule
    \end{tabular}%
    }
    \caption{Comparison of open-weight and fully-open language models. 
    Training FLOPs are estimated using the $6ND$ approximation and reported in units of $10^{21}$.
    Data (T) denotes pretraining tokens in trillions.
    Looped denotes the default number of recurrences used by each model. {\setlength{\fboxsep}{1pt}\colorbox{loopedrow}{\strut Rows in light blue}} refer to looped language models, including our fully-open models. 
    }
    
    \label{tab:scaling_results}
\end{table}

\subsection{Results}

\paragraph{\modelname{} scales to realistic LLM pretraining, with substantial improvements on math and coding over \prenorm.}
Table~\ref{tab:scaling_results} reports our large-scale training results.
Under the same 0.6B parameter scale, 180B-token training budget, and $R=3$ recurrence,
\recurrentSmallModel{} outperforms \prenormSmallModel{} on 7 of 8 evaluated tasks.
The gains are particularly large on mathematical reasoning and code generation,
including +5.2 points on GSM8K and +3.1 points on MBPP.
These results suggest that the benefits of \modelname{} indeed transfer to a realistic LLM pretraining recipe.
Interestingly, while the gains in validation perplexity are relatively modest in
the controlled experiments, they become substantially larger on complex generative tasks.
This highlights the importance of evaluating residual designs beyond perplexity
and multiple-choice benchmarks, which is missing in existing work \citep{prairie2026parcaescalinglawsstable, li2026deeploopdepthscalinglooped, zeitoun2026hyperlooptransformers}. 

\begin{wrapfigure}{r}{0.50\textwidth}
    \centering
    \includegraphics[width=\linewidth]{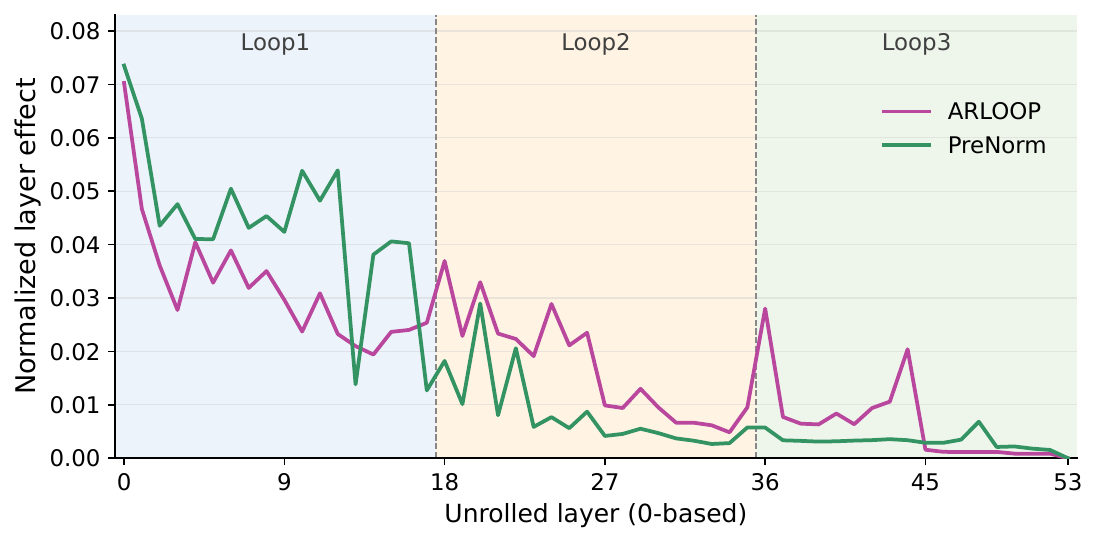}
    \caption{
    Layer effects on 50 MATH examples for \prenormSmallModel{} and \recurrentSmallModel{}. 
    Shaded regions mark Loop1 (layers 0--17), Loop2 (18--35), and Loop3 (36--53).
    }
    \label{fig:math_layer_contribution}
\end{wrapfigure}
\paragraph{Further scaling makes \modelname{} competitive with strong LLMs at a similar scale.}
Scaling \modelname{} to 1.2B parameters and 500B training tokens yields substantial improvements across all evaluated benchmarks.
In particular,
\recurrentLargeModel{} outperforms Huginn on every task in Table~\ref{tab:scaling_results},
despite using fewer parameters, fewer training tokens, and substantially fewer recurrent steps.
Although \recurrentLargeModel{} underperforms Ouro \citep{zhu2025scaling} on most tasks,
we attribute this gap primarily to their substantially larger pretraining data budget and its use of higher-quality pretraining data sources.
Instead, our training recipe relies on open and commonly used data sources.
Our model is also competitive with Qwen3.5-2B despite its substantially smaller pretraining budget,
matching or exceeding it on all evaluated benchmarks.
\recurrentLargeModel{} also outperforms
Gemma-3-4B on both math benchmarks and OLMo models on HumanEval,
demonstrating competitive math and coding capabilities
against substantially larger non-looped LLMs.
These results show that attention residuals can substantially strengthen looped LLMs,
making them competitive with strong non-looped LLMs at similar parameter scales.

\paragraph{Deeper recurrence in \modelname{} contributes more strongly to subsequent computation on math and coding tasks.}
To better understand the gains of \modelname{} on math and coding tasks, we adopt the layer effect metric from \citet{csordas2025languagemodelsusedepth} to quantify how much each layer contributes to representations at subsequent layers.
We randomly sample 50 examples from MATH and, for each example, intervene on the representation of a source layer and measure the resulting change in the outputs of deeper layers.
Figure~\ref{fig:math_layer_contribution} compares the layer effects between \prenormSmallModel{} and \recurrentSmallModel{}.
We find that layers in the second and third recurrences of \modelname{} make substantially larger contributions to subsequent layers than the corresponding deeper layers in \prenormSmallModel{}.
This suggests that \modelname{} makes greater use of representations computed at later recurrent depths, consistent with its stronger performance on tasks that benefit from deeper computation.
We report the detailed score computation and results on MBPP in Appendix~\ref{app:influence_scores}.

\section{Further discussion} \label{sec:discussion}

\paragraph{Dynamic versus fixed recurrence.}
Existing looped language models often vary the number of recurrences during training
to support multiple effective depths
\citep{geiping2025scaling,prairie2026parcaescalinglawsstable, kohli2026loopthinkgeneralize,
bae2025mixtureofrecursionslearningdynamicrecursive}. 
We compare such dynamic-recurrence training with fixed $R=4$ training and find
that fixed recurrence consistently achieves better validation perplexity and
downstream performance at both the 140M and 370M scales.
We therefore use fixed recurrence in our main experiments.
Complete training setups and results are provided in
Appendix~\ref{app:dynamic_recurrence}.

\paragraph{Inference-time scaling.}

We also investigate whether the improved residual access of \modelname{}
enables inference-time recurrence extrapolation, i.e.\ benefiting from more
recurrences than observed during training \citep{fan2024looped,kohli2026loopthinkgeneralize}.
Consistent with prior observations for looped language models
\citep{prairie2026parcaescalinglawsstable,zhu2025scaling},
we find that performance improves up to the maximum training recurrence but
does not further improve beyond it.
This suggests that selective access to earlier representations alone does not
unlock recurrence extrapolation: effectively composing information within the
trained depth range and generalizing recurrent computation beyond that range
appear to be distinct problems. 
Complete experimental details are provided in
Appendix~\ref{app:recurrence_extrapolation}.

\begin{wraptable}{r}{0.5\textwidth}
    \centering
    \scriptsize
    \setlength{\tabcolsep}{3pt}
    \renewcommand{\arraystretch}{1.12}
    \begin{tabular*}{\linewidth}{@{\extracolsep{\fill}}lrrrr@{}}
        \toprule
        Model & 140M & 370M & 0.6B & 1.2B \\
        \midrule
        \multicolumn{5}{c}{Memory (GiB) $\downarrow$} \\
        \midrule
        PreNorm            & 32.8 & 80.0 & 66.5 & 65.4 \\
        \modelnamepl{}  & 33.0 & 80.1 & 66.6 & 65.4 \\
        \modelnameplsw{}   & 32.9 & 80.1 & 66.6 & 65.4 \\
        \midrule
        \multicolumn{5}{c}{Throughput (tokens/s) $\uparrow$} \\
        \midrule
        PreNorm            & 131.1k & 45.8k & 28.9k & 15.0k \\
        \modelnamepl{}  & 107.5k & 39.1k & 24.8k & 13.0k \\
        \modelnameplsw{}  & 114.5k & 40.9k & 25.9k & 13.6k \\
        \bottomrule
    \end{tabular*}
    \caption{Training memory and throughput at $R=4$ on a single NVIDIA H100, with batch sizes of 8, 8, 4, and 2 for the 140M, 370M, 0.6B, and 1.2B models, respectively.
    }
    \label{tab:training_efficiency}
\end{wraptable}
\paragraph{Training efficiency.}
\modelname{} introduces additional computation due to the attention residual calculation for each sublayer.
When activation checkpointing (e.g.\ recomputation of intermediate hidden states) is not enabled, the residual sources used by attention residual are already retained for backpropagation, resulting in negligible additional memory overhead.
We report training memory and throughput for our experimental setups in Table~\ref{tab:training_efficiency}.
Across all model scales, \modelname{} uses essentially the same amount of memory as PreNorm.
In terms of throughput, \modelnamepl{} is approximately 15\% slower than PreNorm on average, while sliding window attention residual reduces this gap to approximately 10\%.
We expect that further kernel optimization could reduce this overhead, as depth attention introduces relatively little additional arithmetic computation but incurs additional high-bandwidth memory traffic.

\section{Conclusion}
We introduce \modelname{}, a looped transformer architecture that adapts attention residuals to recurrent depth.
Through controlled experiments across multiple parameter scales, \modelname{} consistently outperforms PreNorm and prior residual designs for looped transformers.
Distinct from standard transformers, we find that looped transformers do not require access to the full residual history: sliding-window depth attention can match or outperform full-history attention at larger recurrence depths.
More importantly, we demonstrate that these gains transfer to realistic LLM pretraining.
At the 0.6B scale, \modelname{} substantially improves over a matched PreNorm model, with particularly large gains on math and coding tasks.
Our strongest model \recurrentLargeModel{} substantially outperforms the fully-open looped LLM Huginn and is competitive with Qwen3.5-2B despite being trained on substantially less data.
Together, these results demonstrate that selective residual access provides an effective way for looped transformers to better exploit recurrent computation at scale.

\section*{Acknowledgments}

We thank the OSU NLP group for their thoughtful comments. This research was supported in part by NSF CAREER Award \#1942980, Coefficient Giving (formerly Open Philanthropy), gift funding from Amazon, and the Ohio Supercomputer Center \citep{Ohio_Supercomputer_Center1987-dl}.
The views and conclusions expressed herein are those of the authors and do not necessarily reflect the official policies of the U.S. Government, either expressed or implied. The U.S. Government is authorized to reproduce and distribute reprints for government purposes notwithstanding any copyright notice herein.

\bibliography{iclr2027_conference}

@misc{kimi2026attentionresiduals,
  title        = {Attention Residuals},
  author       = {{Kimi Team}},
  year         = {2026},
  eprint       = {2603.15031},
  archivePrefix= {arXiv},
  primaryClass = {cs.CL},
  doi          = {10.48550/arXiv.2603.15031},
  url          = {https://arxiv.org/abs/2603.15031}
}

@article{penedo2024fineweb,
  title   = {The {FineWeb} Datasets: Decanting the Web for the Finest Text Data at Scale},
  author  = {Penedo, Guilherme and Kydl{\'{i}}\v{c}ek, Hynek and Ben Allal, Loubna and Lozhkov, Anton and Mitchell, Margaret and Raffel, Colin and von Werra, Leandro and Wolf, Thomas},
  journal = {arXiv preprint arXiv:2406.17557},
  year    = {2024}
}

@article{hendrycks2021math,
  title   = {Measuring Mathematical Problem Solving With the {MATH} Dataset},
  author  = {Hendrycks, Dan and Burns, Collin and Kadavath, Saurav and Arora, Akul and Basart, Steven and Tang, Eric and Song, Dawn and Steinhardt, Jacob},
  journal = {arXiv preprint arXiv:2103.03874},
  year    = {2021}
}

@article{austin2021program,
  title   = {Program Synthesis with Large Language Models},
  author  = {Austin, Jacob and Odena, Augustus and Nye, Maxwell and Bosma, Maarten and Michalewski, Henryk and Dohan, David and Jiang, Ellen and Cai, Carrie and Terry, Michael and Le, Quoc and Sutton, Charles},
  journal = {arXiv preprint arXiv:2108.07732},
  year    = {2021}
}

@article{chen2021evaluating,
  title   = {Evaluating Large Language Models Trained on Code},
  author  = {Chen, Mark and Tworek, Jerry and Jun, Heewoo and Yuan, Qiming and de Oliveira Pinto, Henrique Ponde and Kaplan, Jared and Edwards, Harri and Burda, Yuri and Joseph, Nicholas and Brockman, Greg and others},
  journal = {arXiv preprint arXiv:2107.03374},
  year    = {2021}
}

@article{cobbe2021training,
  title={Training verifiers to solve math word problems},
  author={Cobbe, Karl and Kosaraju, Vineet and Bavarian, Mohammad and Chen, Mark and Jun, Heewoo and Kaiser, Lukasz and Plappert, Matthias and Tworek, Jerry and Hilton, Jacob and Nakano, Reiichiro and others},
  journal={arXiv preprint arXiv:2110.14168},
  year={2021}
}

@inproceedings{
geiping2025scaling,
title={Scaling up Test-Time Compute with Latent Reasoning: A Recurrent Depth Approach},
author={Jonas Geiping and Sean Michael McLeish and Neel Jain and John Kirchenbauer and Siddharth Singh and Brian R. Bartoldson and Bhavya Kailkhura and Abhinav Bhatele and Tom Goldstein},
booktitle={ES-FoMo III: 3rd Workshop on Efficient Systems for Foundation Models},
year={2025},
url={https://openreview.net/forum?id=D6o6Bwtq7h}
}

@inproceedings{
dehghani2018universal,
title={Universal Transformers},
author={Mostafa Dehghani and Stephan Gouws and Oriol Vinyals and Jakob Uszkoreit and Lukasz Kaiser},
booktitle={International Conference on Learning Representations},
year={2019},
url={https://openreview.net/forum?id=HyzdRiR9Y7},
}

@inproceedings{
Lan2020ALBERT:,
title={ALBERT: A Lite BERT for Self-supervised Learning of Language Representations},
author={Zhenzhong Lan and Mingda Chen and Sebastian Goodman and Kevin Gimpel and Piyush Sharma and Radu Soricut},
booktitle={International Conference on Learning Representations},
year={2020},
url={https://openreview.net/forum?id=H1eA7AEtvS}
}

@article{zhu2025scaling,
  title={Scaling latent reasoning via looped language models},
  author={Zhu, Rui-Jie and Wang, Zixuan and Hua, Kai and Zhang, Tianyu and Li, Ziniu and Que, Haoran and Wei, Boyi and Wen, Zixin and Yin, Fan and Xing, He and others},
  journal={arXiv preprint arXiv:2510.25741},
  year={2025}
}

@article{qwen2025qwen3,
  title={Qwen3 Technical Report},
  author={{Qwen Team}},
  journal={arXiv preprint arXiv:2505.09388},
  year={2025}
}

@article{gemmateam2025gemma3,
  title={Gemma 3 Technical Report},
  author={{Gemma Team}},
  journal={arXiv preprint arXiv:2503.19786},
  year={2025}
}

@misc{huggingfacetb2025smollm3,
  title={{SmolLM3-3B-Base} Model Card},
  author={{HuggingFaceTB}},
  year={2025},
  url={https://huggingface.co/HuggingFaceTB/SmolLM3-3B-Base}
}

@misc{qwen2026qwen35base,
  title={{Qwen3.5-2B-Base} Model Card},
  author={{Qwen Team}},
  year={2026},
  url={https://huggingface.co/Qwen/Qwen3.5-2B-Base}
}

@article{wang2026hrmtext,
  title={{HRM-Text}: Efficient Pretraining Beyond Scaling},
  author={Wang, Guan and Liu, Changling and Wang, Chenyu and Zhou, Cai and Sun, Yuhao and Wu, Yifei and Zhen, Shuai and Scimeca, Luca and Abbasi-Yadkori, Yasin},
  journal={arXiv preprint arXiv:2605.20613},
  year={2026}
}

@misc{allenai2025olmo3base,
  title={{Olmo-3-1025-7B} Model Card},
  author={{Allen Institute for AI}},
  year={2025},
  url={https://huggingface.co/allenai/Olmo-3-1025-7B}
}

@article{olmo2024furious,
  title={{2 OLMo 2 Furious}},
  author={{Team OLMo} and Walsh, Pete and Soldaini, Luca and Groeneveld, Dirk and Lo, Kyle and Arora, Shane and Bhagia, Akshita and Gu, Yuling and Huang, Shengyi and Jordan, Matt and others},
  journal={arXiv preprint arXiv:2501.00656},
  year={2024}
}

@article{allal2025smollm2,
  title={SmolLM2: When Smol Goes Big---Data-Centric Training of a Small Language Model},
  author={Allal, Loubna Ben and Lozhkov, Anton and Bakouch, Elie and Bl{\'a}zquez, Gabriel Mart{\'i}n and Penedo, Guilherme and Tunstall, Lewis and Marafioti, Andr{\'e}s and Kydl{\'i}{\v c}ek, Hynek and Lajar{\'i}n, Agust{\'i}n Piqueres and Srivastav, Vaibhav and others},
  journal={arXiv preprint arXiv:2502.02737},
  year={2025}
}

@inproceedings{
yang2024looped,
title={Looped Transformers are Better at Learning Learning Algorithms},
author={Liu Yang and Kangwook Lee and Robert D Nowak and Dimitris Papailiopoulos},
booktitle={The Twelfth International Conference on Learning Representations},
year={2024},
url={https://openreview.net/forum?id=HHbRxoDTxE}
}

@inproceedings{
fan2024looped,
title={Looped Transformers for Length Generalization},
author={Ying Fan and Yilun Du and Kannan Ramchandran and Kangwook Lee},
booktitle={The Thirteenth International Conference on Learning Representations},
year={2025},
url={https://openreview.net/forum?id=2edigk8yoU}
}

@inproceedings{
saunshireasoning,
title={Reasoning with Latent Thoughts: On the Power of Looped Transformers},
author={Nikunj Saunshi and Nishanth Dikkala and Zhiyuan Li and Sanjiv Kumar and Sashank J. Reddi},
booktitle={The Thirteenth International Conference on Learning Representations},
year={2025},
url={https://openreview.net/forum?id=din0lGfZFd}
}

@inproceedings{NIPS2017_3f5ee243,
 author = {Vaswani, Ashish and Shazeer, Noam and Parmar, Niki and Uszkoreit, Jakob and Jones, Llion and Gomez, Aidan N and Kaiser, \L ukasz and Polosukhin, Illia},
 booktitle = {Advances in Neural Information Processing Systems},
 editor = {I. Guyon and U. Von Luxburg and S. Bengio and H. Wallach and R. Fergus and S. Vishwanathan and R. Garnett},
 pages = {},
 publisher = {Curran Associates, Inc.},
 title = {Attention is All you Need},
 url = {https://proceedings.neurips.cc/paper_files/paper/2017/file/3f5ee243547dee91fbd053c1c4a845aa-Paper.pdf},
 volume = {30},
 year = {2017}
}

@article{bansal2022end,
  title={End-to-end algorithm synthesis with recurrent networks: Extrapolation without overthinking},
  author={Bansal, Arpit and Schwarzschild, Avi and Borgnia, Eitan and Emam, Zeyad and Huang, Furong and Goldblum, Micah and Goldstein, Tom},
  journal={Advances in Neural Information Processing Systems},
  volume={35},
  pages={20232--20242},
  year={2022}
}

@article{radford2019language,
  title={Language models are unsupervised multitask learners},
  author={Radford, Alec and Wu, Jeffrey and Child, Rewon and Luan, David and Amodei, Dario and Sutskever, Ilya and others},
  journal={OpenAI blog},
  volume={1},
  number={8},
  pages={9},
  year={2019}
}

@inproceedings{
csordas2024moeut,
title={Mo{EUT}: Mixture-of-Experts Universal Transformers},
author={R{\'o}bert Csord{\'a}s and Kazuki Irie and J{\"u}rgen Schmidhuber and Christopher Potts and Christopher D Manning},
booktitle={The Thirty-eighth Annual Conference on Neural Information Processing Systems},
year={2024},
url={https://openreview.net/forum?id=ZxVrkm7Bjl}
}

@MISC{Ohio_Supercomputer_Center1987-dl,
 title   = "Ohio Supercomputer Center",
 author  = "{Ohio Supercomputer Center}",
 publisher = "Ohio Supercomputer Center",
 year   = "1987",
 url    = "https://ror.org/01apna436"
}

@misc{prairie2026parcaescalinglawsstable,
      title={Parcae: Scaling Laws For Stable Looped Language Models}, 
      author={Hayden Prairie and Zachary Novack and Taylor Berg-Kirkpatrick and Daniel Y. Fu},
      year={2026},
      eprint={2604.12946},
      archivePrefix={arXiv},
      primaryClass={cs.LG},
      url={https://arxiv.org/abs/2604.12946}, 
}

@misc{li2026deeploopdepthscalinglooped,
      title={DeepLoop: Depth Scaling for Looped Transformers}, 
      author={Shuzhen Li and Yifan Zhang and Jiacheng Guo and Quanquan Gu and Mengdi Wang},
      year={2026},
      eprint={2607.13491},
      archivePrefix={arXiv},
      primaryClass={cs.LG},
      url={https://arxiv.org/abs/2607.13491}, 
}

@misc{gao2026looploopies,
      title={Loop the Loopies!}, 
      author={Zitian Gao and Yilong Chen and Yihao Xiao and Xinyu Yang and Ran Tao and Joey Zhou and Bryan Dai},
      year={2026},
      eprint={2607.16051},
      archivePrefix={arXiv},
      primaryClass={cs.CL},
      url={https://arxiv.org/abs/2607.16051}, 
}

@misc{wang2026smeltscalinglawscomputematched,
      title={SMELT: Scaling Laws for Compute-Matched MoE Looped Transformers}, 
      author={Shaowen Wang and Ge Zhang and Kairong Luo and Yuhao Wu and Shaofan Liu and Jiaheng Liu and Wenhao Huang and Shen Yan and Jian Li},
      year={2026},
      eprint={2609.01343},
      archivePrefix={arXiv},
      primaryClass={cs.LG},
      url={https://arxiv.org/abs/2609.01343}, 
}

@misc{nanochat,
  author = {Andrej Karpathy},
  title = {nanochat: The best ChatGPT that \$100 can buy},
  year = {2025},
  publisher = {GitHub},
  url = {https://github.com/karpathy/nanochat}
}

@misc{xiong2020layernormalizationtransformerarchitecture,
      title={On Layer Normalization in the Transformer Architecture}, 
      author={Ruibin Xiong and Yunchang Yang and Di He and Kai Zheng and Shuxin Zheng and Chen Xing and Huishuai Zhang and Yanyan Lan and Liwei Wang and Tie-Yan Liu},
      year={2020},
      eprint={2002.04745},
      archivePrefix={arXiv},
      primaryClass={cs.LG},
      url={https://arxiv.org/abs/2002.04745}, 
}

@misc{zeitoun2026hyperlooptransformers,
      title={Hyperloop Transformers}, 
      author={Abbas Zeitoun and Lucas Torroba-Hennigen and Yoon Kim},
      year={2026},
      eprint={2604.21254},
      archivePrefix={arXiv},
      primaryClass={cs.LG},
      url={https://arxiv.org/abs/2604.21254}, 
}

@misc{kohli2026loopthinkgeneralize,
      title={Loop, Think, \& Generalize: Implicit Reasoning in Recurrent-Depth Transformers},
      author={Harsh Kohli and Srinivasan Parthasarathy and Huan Sun and Yuekun Yao},
      year={2026},
      eprint={2604.07822},
      archivePrefix={arXiv},
      primaryClass={cs.CL},
      url={https://arxiv.org/abs/2604.07822}, 
}

@misc{csordas2025languagemodelsusedepth,
      title={Do Language Models Use Their Depth Efficiently?}, 
      author={Róbert Csordás and Christopher D. Manning and Christopher Potts},
      year={2025},
      eprint={2505.13898},
      archivePrefix={arXiv},
      primaryClass={cs.LG},
      url={https://arxiv.org/abs/2505.13898}, 
}

@misc{ainslie2023gqatraininggeneralizedmultiquery,
      title={GQA: Training Generalized Multi-Query Transformer Models from Multi-Head Checkpoints}, 
      author={Joshua Ainslie and James Lee-Thorp and Michiel de Jong and Yury Zemlyanskiy and Federico Lebrón and Sumit Sanghai},
      year={2023},
      eprint={2305.13245},
      archivePrefix={arXiv},
      primaryClass={cs.CL},
      url={https://arxiv.org/abs/2305.13245}, 
}

@misc{shazeer2020gluvariantsimprovetransformer,
      title={GLU Variants Improve Transformer}, 
      author={Noam Shazeer},
      year={2020},
      eprint={2002.05202},
      archivePrefix={arXiv},
      primaryClass={cs.LG},
      url={https://arxiv.org/abs/2002.05202}, 
}

@misc{clark2018thinksolvedquestionanswering,
      title={Think you have Solved Question Answering? Try ARC, the AI2 Reasoning Challenge}, 
      author={Peter Clark and Isaac Cowhey and Oren Etzioni and Tushar Khot and Ashish Sabharwal and Carissa Schoenick and Oyvind Tafjord},
      year={2018},
      eprint={1803.05457},
      archivePrefix={arXiv},
      primaryClass={cs.AI},
      url={https://arxiv.org/abs/1803.05457}, 
}

@misc{zellers2019hellaswagmachinereallyfinish,
      title={HellaSwag: Can a Machine Really Finish Your Sentence?}, 
      author={Rowan Zellers and Ari Holtzman and Yonatan Bisk and Ali Farhadi and Yejin Choi},
      year={2019},
      eprint={1905.07830},
      archivePrefix={arXiv},
      primaryClass={cs.CL},
      url={https://arxiv.org/abs/1905.07830}, 
}

@misc{sakaguchi2019winograndeadversarialwinogradschema,
      title={WinoGrande: An Adversarial Winograd Schema Challenge at Scale}, 
      author={Keisuke Sakaguchi and Ronan Le Bras and Chandra Bhagavatula and Yejin Choi},
      year={2019},
      eprint={1907.10641},
      archivePrefix={arXiv},
      primaryClass={cs.CL},
      url={https://arxiv.org/abs/1907.10641}, 
}

@misc{mihaylov2018suitarmorconductelectricity,
      title={Can a Suit of Armor Conduct Electricity? A New Dataset for Open Book Question Answering}, 
      author={Todor Mihaylov and Peter Clark and Tushar Khot and Ashish Sabharwal},
      year={2018},
      eprint={1809.02789},
      archivePrefix={arXiv},
      primaryClass={cs.CL},
      url={https://arxiv.org/abs/1809.02789}, 
}

@misc{lewkowycz2022solvingquantitativereasoningproblems,
      title={Solving Quantitative Reasoning Problems with Language Models}, 
      author={Aitor Lewkowycz and Anders Andreassen and David Dohan and Ethan Dyer and Henryk Michalewski and Vinay Ramasesh and Ambrose Slone and Cem Anil and Imanol Schlag and Theo Gutman-Solo and Yuhuai Wu and Behnam Neyshabur and Guy Gur-Ari and Vedant Misra},
      year={2022},
      eprint={2206.14858},
      archivePrefix={arXiv},
      primaryClass={cs.CL},
      url={https://arxiv.org/abs/2206.14858}, 
}

@misc{bae2025mixtureofrecursionslearningdynamicrecursive,
      title={Mixture-of-Recursions: Learning Dynamic Recursive Depths for Adaptive Token-Level Computation}, 
      author={Sangmin Bae and Yujin Kim and Reza Bayat and Sungnyun Kim and Jiyoun Ha and Tal Schuster and Adam Fisch and Hrayr Harutyunyan and Ziwei Ji and Aaron Courville and Se-Young Yun},
      year={2025},
      eprint={2507.10524},
      archivePrefix={arXiv},
      primaryClass={cs.CL},
      url={https://arxiv.org/abs/2507.10524}, 
}

@inproceedings{paperno-etal-2016-lambada,
    title = "The {LAMBADA} dataset: Word prediction requiring a broad discourse context",
    author = "Paperno, Denis  and
      Kruszewski, Germ{\'a}n  and
      Lazaridou, Angeliki  and
      Pham, Ngoc Quan  and
      Bernardi, Raffaella  and
      Pezzelle, Sandro  and
      Baroni, Marco  and
      Boleda, Gemma  and
      Fern{\'a}ndez, Raquel",
    editor = "Erk, Katrin  and
      Smith, Noah A.",
    booktitle = "Proceedings of the 54th Annual Meeting of the Association for Computational Linguistics (Volume 1: Long Papers)",
    month = aug,
    year = "2016",
    address = "Berlin, Germany",
    publisher = "Association for Computational Linguistics",
    url = "https://aclanthology.org/P16-1144/",
    doi = "10.18653/v1/P16-1144",
    pages = "1525--1534"
}

@misc{bisk2019piqareasoningphysicalcommonsense,
      title={PIQA: Reasoning about Physical Commonsense in Natural Language}, 
      author={Yonatan Bisk and Rowan Zellers and Ronan Le Bras and Jianfeng Gao and Yejin Choi},
      year={2019},
      eprint={1911.11641},
      archivePrefix={arXiv},
      primaryClass={cs.CL},
      url={https://arxiv.org/abs/1911.11641}, 
}

@inproceedings{welbl-etal-2017-crowdsourcing,
    title = "Crowdsourcing Multiple Choice Science Questions",
    author = "Welbl, Johannes  and
      Liu, Nelson F.  and
      Gardner, Matt",
    editor = "Derczynski, Leon  and
      Xu, Wei  and
      Ritter, Alan  and
      Baldwin, Tim",
    booktitle = "Proceedings of the 3rd Workshop on Noisy User-generated Text",
    month = sep,
    year = "2017",
    address = "Copenhagen, Denmark",
    publisher = "Association for Computational Linguistics",
    url = "https://aclanthology.org/W17-4413/",
    doi = "10.18653/v1/W17-4413",
    pages = "94--106"
}

@inproceedings{lai-etal-2017-race,
    title = "{RACE}: Large-scale {R}e{A}ding Comprehension Dataset From Examinations",
    author = "Lai, Guokun  and
      Xie, Qizhe  and
      Liu, Hanxiao  and
      Yang, Yiming  and
      Hovy, Eduard",
    editor = "Palmer, Martha  and
      Hwa, Rebecca  and
      Riedel, Sebastian",
    booktitle = "Proceedings of the 2017 Conference on Empirical Methods in Natural Language Processing",
    month = sep,
    year = "2017",
    address = "Copenhagen, Denmark",
    publisher = "Association for Computational Linguistics",
    url = "https://aclanthology.org/D17-1082/",
    doi = "10.18653/v1/D17-1082",
    pages = "785--794"
}

@misc{kingma2017adammethodstochasticoptimization,
      title={Adam: A Method for Stochastic Optimization}, 
      author={Diederik P. Kingma and Jimmy Ba},
      year={2017},
      eprint={1412.6980},
      archivePrefix={arXiv},
      primaryClass={cs.LG},
      url={https://arxiv.org/abs/1412.6980}, 
}

@misc{zhou2025valueresiduallearning,
      title={Value Residual Learning}, 
      author={Zhanchao Zhou and Tianyi Wu and Zhiyun Jiang and Fares Obeid and Zhenzhong Lan},
      year={2025},
      eprint={2410.17897},
      archivePrefix={arXiv},
      primaryClass={cs.CL},
      url={https://arxiv.org/abs/2410.17897}, 
}

@inproceedings{sennrich-etal-2016-neural,
    title = "Neural Machine Translation of Rare Words with Subword Units",
    author = "Sennrich, Rico  and
      Haddow, Barry  and
      Birch, Alexandra",
    editor = "Erk, Katrin  and
      Smith, Noah A.",
    booktitle = "Proceedings of the 54th Annual Meeting of the Association for Computational Linguistics (Volume 1: Long Papers)",
    month = aug,
    year = "2016",
    address = "Berlin, Germany",
    publisher = "Association for Computational Linguistics",
    url = "https://aclanthology.org/P16-1162/",
    doi = "10.18653/v1/P16-1162",
    pages = "1715--1725"
}

@misc{eval-harness,
  author       = {Gao, Leo and Tow, Jonathan and Abbasi, Baber and Biderman, Stella and Black, Sid and DiPofi, Anthony and Foster, Charles and Golding, Laurence and Hsu, Jeffrey and Le Noac'h, Alain and Li, Haonan and McDonell, Kyle and Muennighoff, Niklas and Ociepa, Chris and Phang, Jason and Reynolds, Laria and Schoelkopf, Hailey and Skowron, Aviya and Sutawika, Lintang and Tang, Eric and Thite, Anish and Wang, Ben and Wang, Kevin and Zou, Andy},
  title        = {The Language Model Evaluation Harness},
  month        = 07,
  year         = 2024,
  publisher    = {Zenodo},
  version      = {v0.4.3},
  doi          = {10.5281/zenodo.12608602},
  url          = {https://zenodo.org/records/12608602}
}

@article{ahmad2025opencodereasoning,
      title={OpenCodeReasoning: Advancing Data Distillation for Competitive Coding}, 
      author={Wasi Uddin Ahmad and Sean Narenthiran and Somshubra Majumdar and Aleksander Ficek and Siddhartha Jain and Jocelyn Huang and Vahid Noroozi and Boris Ginsburg},
      year={2025},
      eprint={2504.01943},
      archivePrefix={arXiv},
      primaryClass={cs.CL},
      url={https://arxiv.org/abs/2504.01943}, 
}

@misc{liu2023codegeneratedchatgptreally,
      title={Is Your Code Generated by ChatGPT Really Correct? Rigorous Evaluation of Large Language Models for Code Generation}, 
      author={Jiawei Liu and Chunqiu Steven Xia and Yuyao Wang and Lingming Zhang},
      year={2023},
      eprint={2305.01210},
      archivePrefix={arXiv},
      primaryClass={cs.SE},
      url={https://arxiv.org/abs/2305.01210}, 
}

@misc{lozhkov2024starcoder,
      title={StarCoder 2 and The Stack v2: The Next Generation}, 
      author={Anton Lozhkov and Raymond Li and Loubna Ben Allal and Federico Cassano and Joel Lamy-Poirier and Nouamane Tazi and Ao Tang and Dmytro Pykhtar and Jiawei Liu and Yuxiang Wei and Tianyang Liu and Max Tian and Denis Kocetkov and Arthur Zucker and Younes Belkada and Zijian Wang and Qian Liu and Dmitry Abulkhanov and Indraneil Paul and Zhuang Li and Wen-Ding Li and Megan Risdal and Jia Li and Jian Zhu and Terry Yue Zhuo and Evgenii Zheltonozhskii and Nii Osae Osae Dade and Wenhao Yu and Lucas Krauß and Naman Jain and Yixuan Su and Xuanli He and Manan Dey and Edoardo Abati and Yekun Chai and Niklas Muennighoff and Xiangru Tang and Muhtasham Oblokulov and Christopher Akiki and Marc Marone and Chenghao Mou and Mayank Mishra and Alex Gu and Binyuan Hui and Tri Dao and Armel Zebaze and Olivier Dehaene and Nicolas Patry and Canwen Xu and Julian McAuley and Han Hu and Torsten Scholak and Sebastien Paquet and Jennifer Robinson and Carolyn Jane Anderson and Nicolas Chapados and Mostofa Patwary and Nima Tajbakhsh and Yacine Jernite and Carlos Muñoz Ferrandis and Lingming Zhang and Sean Hughes and Thomas Wolf and Arjun Guha and Leandro von Werra and Harm de Vries},
      year={2024},
      eprint={2402.19173},
      archivePrefix={arXiv},
      primaryClass={cs.SE}
}

@misc{yuan2025naturalreasoningreasoningwild28m,
      title={NaturalReasoning: Reasoning in the Wild with 2.8M Challenging Questions}, 
      author={Weizhe Yuan and Jane Yu and Song Jiang and Karthik Padthe and Yang Li and Dong Wang and Ilia Kulikov and Kyunghyun Cho and Yuandong Tian and Jason E Weston and Xian Li},
      year={2025},
      eprint={2502.13124},
      archivePrefix={arXiv},
      primaryClass={cs.CL},
      url={https://arxiv.org/abs/2502.13124}, 
}

@article{toshniwal2024openmath2,
  title   = {OpenMathInstruct-2: Accelerating AI for Math with Massive Open-Source Instruction Data},
  author  = {Shubham Toshniwal and Wei Du and Ivan Moshkov and  Branislav Kisacanin and Alexan Ayrapetyan and Igor Gitman},
  year    = {2024},
  journal = {arXiv preprint arXiv:2410.01560}
}

@article{moshkov2025aimo2,
  title   = {AIMO-2 Winning Solution: Building State-of-the-Art Mathematical Reasoning Models with OpenMathReasoning dataset},
  author  = {Ivan Moshkov and Darragh Hanley and Ivan Sorokin and Shubham Toshniwal and Christof Henkel and Benedikt Schifferer and Wei Du and Igor Gitman},
  year    = {2025},
  journal = {arXiv preprint arXiv:2504.16891}
}

@misc{knupp2026depthrecurrentattentionmixturesgiving,
      title={Depth-Recurrent Attention Mixtures: Giving Latent Reasoning the Attention it Deserves}, 
      author={Jonas Knupp and Jan Hendrik Metzen and Jeremias Bohn and Georg Groh and Kristian Kersting},
      year={2026},
      eprint={2601.21582},
      archivePrefix={arXiv},
      primaryClass={cs.AI},
      url={https://arxiv.org/abs/2601.21582}, 
}

@misc{wang2022deepnetscalingtransformers1000,
      title={DeepNet: Scaling Transformers to 1,000 Layers}, 
      author={Hongyu Wang and Shuming Ma and Li Dong and Shaohan Huang and Dongdong Zhang and Furu Wei},
      year={2022},
      eprint={2203.00555},
      archivePrefix={arXiv},
      primaryClass={cs.CL},
      url={https://arxiv.org/abs/2203.00555}, 
}

@misc{huang2026loopedmodelsrightii,
      title={Towards Looped Models Done Right, Part II: Rethinking at Fixed Points}, 
      author={Benhao Huang and Chufan Shi and Junlin Chen and Shicheng Wen and Zhengzhong Liu and Eric Xing and Xuezhe Ma},
      year={2026},
      eprint={2610.06833},
      archivePrefix={arXiv},
      primaryClass={cs.LG},
      url={https://arxiv.org/abs/2610.06833}, 
}

@misc{lab2026nanbeige423bunlockingagenticcapabilities,
      title={Nanbeige4.2-3B: Unlocking Agentic Capabilities in a Compact Model}, 
      author={Nanbeige Lab and : and Chen Yang and Chengrui Huang and Fufeng Lan and Hanhui Chen and Hao Zhou and Huatong Song and Jiaqi Cao and Jiaying Zhu and Jinlin Niu and Kai Wang and Lisheng Huang and Qiliang Liang and Ran Le and Ruixiang Feng and Shuang Sun and Tao Gu and Tao Zhang and Tianyu Luo and Yang Song and Yun Xing and Yuntao Wen and Ziyao Xu and Zongchao Chen and Zongqiang Li},
      year={2026},
      eprint={2607.22083},
      archivePrefix={arXiv},
      primaryClass={cs.AI},
      url={https://arxiv.org/abs/2607.22083}, 
}

@article{yang2026loopcoder,
  title={LoopCoder-v2: Only Loop Once for Efficient Test-Time Computation Scaling},
  author={Yang, Jian and Guo, Shawn and Zhang, Wei and Zheng, Tianyu and Du, Yaxin and Li, Haau-Sing and Wu, Jiajun and Song, Yue and Xing, Yan and Cai, Qingsong and others},
  journal={arXiv preprint arXiv:2606.18023},
  year={2026}
}
\bibliographystyle{iclr2027_conference}

\appendix
\section{Controlled experiment details}

\subsection{Architecture and Training.}
\label{app:exp1_setup}
Table~\ref{tab:controlled_architecture} specifies the 140M and 370M model families used in Section~\ref{sec:exp1}. 
The table describes the shared backbone of PreNorm, \modelnamepl{}, and \modelnameplsw{}. Each recurrence reuses the entire physical layer stack; increasing $R$ increases the effective depth without adding Transformer layers. 
Depth-attention queries and normalization parameters are specific to each effective sublayer.
At the final effective depth, we apply a separate attention-residual aggregation to produce the representation passed to the language modeling head.
We use a scaled-zero initialization, which is the same strategy as \citet{prairie2026parcaescalinglawsstable}.
The additional baselines retain the same model dimensions and use their respective residual updates and initialization rules.

\paragraph{Value residuals.}
Following the default setup of \citet{nanochat}, we use value residuals \citep{zhou2025valueresiduallearning} in all experiments.
Specifically, at alternating effective layers (e.g., the 1st, 3rd, and 5th layers), we inject a gated token embedding into the value vectors of the self-attention sublayer.
These token-specific embedding parameters are distinct from the input embeddings.
Value residual learning has been shown to improve LLM pretraining \citep{zhou2025valueresiduallearning} and has also been adopted in recent looped-transformer architectures \citep{prairie2026parcaescalinglawsstable}.

Although value residuals are not a contribution of our work, we find them important for obtaining strong looped-transformer performance, particularly at smaller scales.
Importantly, removing value residuals substantially degrades our sliding-window depth-attention variant.
They are even more important for Sandwich-Norm looped transformers: without value residuals, training becomes unstable and the loss remains near 7.5 rather than continuing to decrease.
This pattern is consistent with the failed Sandwich-Norm training run reported by \citet[Figure~5, Bad Run 1]{geiping2025scaling}.
We hypothesize that this instability is related to representation collapse at large effective depths, where representations of different tokens become increasingly similar, and
repeated injection of token-specific information through value residuals mitigates this effect.
Empirically, adding value residuals eliminates the observed training failure and makes the Sandwich-Norm variant trainable in our setting.
\begin{table}[t]
    \centering
    \scriptsize
    \setlength{\tabcolsep}{4pt}
    \renewcommand{\arraystretch}{1.08}
    \begin{tabular}{@{}llcccc@{}}
        \toprule
        & & \multicolumn{2}{c}{140M} & \multicolumn{2}{c}{370M} \\
        \cmidrule(lr){3-4}\cmidrule(lr){5-6}
        & & PreNorm & \modelname{} & PreNorm & \modelname{} \\
        \midrule
        \multirow{5}{*}{\rotatebox{90}{\emph{Architecture}}}
        & Physical Transformer layers $L$ & 6 & 6 & 12 & 12 \\
        & Hidden dimension $d_{\mathrm{model}}$ & 768 & 768 & 1,024 & 1,024 \\
        & MLP dimension $d_{\mathrm{intermediate}}$ & 3,072 & 3,072 & 4,096 & 4,096 \\
        & Attention heads (query / key--value) & 6 / 6 & 6 / 6 & 8 / 8 & 8 / 8 \\
        & Head dimension & 128 & 128 & 128 & 128 \\
        \midrule
        \multirow{10}{*}{\rotatebox{90}{\emph{Shared details}}}
        & Attention & \multicolumn{4}{c}{Causal multi-head attention with QK-Norm} \\
        & MLP activation & \multicolumn{4}{c}{ReLU$^2$} \\
        & Position embeddings & \multicolumn{4}{c}{RoPE ($\theta=50{,}000$)} \\
        & Vocabulary size & \multicolumn{4}{c}{32,768} \\
        & Normalization & \multicolumn{4}{c}{RMSNorm ($\epsilon=10^{-5}$)} \\
        & Context length & \multicolumn{4}{c}{2,048} \\
        & Linear biases & \multicolumn{4}{c}{None} \\
        & Tied input/output embeddings & \multicolumn{4}{c}{Yes} \\
        \midrule
        \multirow{5}{*}{\rotatebox{90}{\emph{Recurrence}}}
        & Training / evaluation recurrence $R$ & \multicolumn{4}{c}{Fixed, $R\in\{1,2,3,4,5\}$ (depending on comparison)} \\
        & Effective Transformer depth $LR$ & $6R$ & $6R$ & $12R$ & $12R$ \\
        & Residual block size (sublayers) & -- & 6 & -- & 12 \\
        & Transformer layers per AttnRes block & -- & 3 & -- & 6 \\
        & Sliding window $W$ & -- & 3 + embedding & -- & 3 + embedding \\
        \bottomrule
    \end{tabular}
    \caption{Architecture details for the controlled 140M and 370M experiments. Each Transformer layer contains an attention sublayer and an MLP sublayer. The AttnRes block and window settings apply to the corresponding residual variants; Table~\ref{tab:ablation} lists the evaluated configurations. PreNorm uses additive pre-normalized residual updates, while \modelname{} learns the input mixture at each effective sublayer.}
    \label{tab:controlled_architecture}
\end{table}

\paragraph{Training configuration.}
We train on 11.2B tokens at 140M and 29.6B tokens at 370M, using a 32,768 vocabulary and packed FineWeb-Edu data. 
We train the vocabulary using BPE \citep{sennrich-etal-2016-neural} on FineWeb-Edu data, following \citet{prairie2026parcaescalinglawsstable}.
The global batch size is 256 sequences of length 2,048, or 524,288 tokens per optimizer update. We use AdamW with $\beta_1=0.9$, $\beta_2=0.95$, $\epsilon=10^{-8}$, weight decay 0.1, and gradient clipping at 1.0. The cosine learning-rate schedule has 1,000 warmup updates. 
Runs use BF16 mixed precision.
Validation uses a fixed set of 4,096 sequences at the model's training recurrence, which is approximately 10M tokens. 

When comparing against residual designs from related work, we empirically observe that Sandwich-Norm requires a lower learning rate than PreNorm. We therefore sweep learning rates over
$\{8\times10^{-4}, 1\times10^{-3}, 2\times10^{-3}, 4\times10^{-3}, 6\times10^{-3}, 8\times10^{-3}\}$
and select the best learning rate separately for PreNorm and Sandwich-Norm \citep{geiping2025scaling}.
For the 140M models, we use a learning rate of $6\times10^{-3}$ for PreNorm and $2\times10^{-3}$ for Sandwich-Norm. For the 370M models, we use $4\times10^{-3}$ for PreNorm and $1\times10^{-3}$ for Sandwich-Norm.
Parcae \citep{prairie2026parcaescalinglawsstable} and our attention residual methods use the same learning rate as PreNorm to enable comparison against the strongest baseline.
DeepLoop \citep{li2026deeploopdepthscalinglooped} uses the same learning rate as Sandwich-Norm, as both methods apply residual normalization and empirically favor lower learning rates.

\subsection{evaluation setup}
\label{app:controlled_evaluation}
Table~\ref{tab:controlled_evaluation_setup} specifies the metrics used in Section~\ref{sec:exp1}. For downstream evaluation, we use the language model evaluation harness \citep{eval-harness} with the same task definitions across residual designs. 
Normalized accuracy divides candidate log-likelihoods by the task's length normalization before selecting an answer. 
\begin{table}[t]
\centering
\small
\setlength{\tabcolsep}{7pt}
\begin{tabular}{@{}lll@{}}
\toprule
\textbf{Benchmark} & \textbf{Settings / reported metric} & \textbf{Framework} \\
\midrule
\multicolumn{3}{c}{\textbf{Language modeling}} \\
FineWeb-Edu (held out) & PPL &  \\
\midrule
\multicolumn{3}{c}{\textbf{Zero-shot downstream evaluation}} \\
ARC-Challenge & Log-likelihood, 0-shot; accuracy & lm-eval-harness \\
ARC-Easy & Log-likelihood, 0-shot; accuracy & lm-eval-harness \\
HellaSwag & Log-likelihood, 0-shot; normalized accuracy & lm-eval-harness \\
LAMBADA (OpenAI) & Last-word prediction, 0-shot; accuracy & lm-eval-harness \\
LAMBADA (standard) & Last-word prediction, 0-shot; accuracy & lm-eval-harness \\
OpenBookQA & Log-likelihood, 0-shot; normalized accuracy & lm-eval-harness \\
PIQA & Log-likelihood, 0-shot; normalized accuracy & lm-eval-harness \\
RACE & Log-likelihood, 0-shot; accuracy & lm-eval-harness \\
SciQ & Log-likelihood, 0-shot; accuracy & lm-eval-harness \\
WinoGrande & Log-likelihood, 0-shot; accuracy & lm-eval-harness \\
\bottomrule
\end{tabular}
\caption{Evaluation protocol for the controlled 140M and 370M experiments. 
}
\label{tab:controlled_evaluation_setup}
\end{table}

\FloatBarrier

\subsection{Ablation without value residuals}
\label{app:no_value_residuals}
Table~\ref{tab:no_value_residuals} reports results for the 140M, $R=3$ setting with gated value embeddings disabled.
The corresponding results with value residuals enabled are reported in Table~\ref{tab:related-model-val-ppl}.
\textit{Collapse} indicates that the training loss fails to decrease, with token representations becoming nearly identical, due to residual normalization.
Even without value residuals, \modelname{} continues to outperform prior residual designs, showing that its gains are not attributable to the use of value residuals.

\begin{table}[!ht]
    \centering
    \small
    \begin{tabular}{@{}lr@{}}
        \toprule
        Model & 140M, $R=3$, w/o value res. (Val PPL $\downarrow$) \\
        \midrule
        PreNorm & 18.19 \\
        Sandwich-Norm \citep{geiping2025scaling} & Collapse \\
        DeepLoop \citep{li2026deeploopdepthscalinglooped} & 18.44 \\
        Parcae + input injection \citep{prairie2026parcaescalinglawsstable} & 17.83 \\
        \modelnamepl{} & \textbf{17.52} \\
        \modelnameplsw{} & 18.20 \\
        \bottomrule
    \end{tabular}
    \caption{Validation perplexity without value residuals. 
    Bold marks the best available result.
    }
    \label{tab:no_value_residuals}
\end{table}
\FloatBarrier

\subsection{Per-task results for controlled pretraining}
\label{app:controlled_task_accuracy}

Table~\ref{tab:controlled_task_accuracy} reports the individual task accuracies underlying the averages in Table~\ref{tab:ablation}, grouped by model scale.

\begin{table}[!ht]
\centering
\scriptsize
\setlength{\tabcolsep}{2.3pt}
\resizebox{\linewidth}{!}{%
\begin{tabular}{@{}clrrrrrrrrrrr@{}}
\toprule
$R$ & Residual design & ARC-C & ARC-E & HellaSwag & \shortstack{LAMBADA\\OpenAI} & \shortstack{LAMBADA\\Standard} & OBQA & PIQA & RACE & SciQ & WinoGrande & Avg. \\
\midrule
\multicolumn{13}{c}{\textbf{140M}} \\
\midrule
\multirow{2}{*}{1} & PreNorm & 21.84 & \textbf{56.36} & \textbf{34.49} & \textbf{25.77} & \textbf{19.13} & \textbf{32.20} & 63.49 & 28.33 & 82.40 & 50.04 & 41.41 \\
 & \modelnamepl{} & \textbf{23.72} & 55.22 & 33.90 & 25.58 & 18.18 & \textbf{32.20} & \textbf{64.15} & \textbf{28.90} & \textbf{83.90} & \textbf{52.49} & \textbf{41.82} \\
\midrule
\multirow{3}{*}{2} & PreNorm & 23.55 & 56.40 & 35.40 & 26.97 & \textbf{21.56} & 28.40 & \textbf{63.93} & \textbf{32.34} & 82.30 & \textbf{51.38} & 42.22 \\
 & \modelnamepl{} & \textbf{23.63} & 55.51 & 35.64 & \textbf{27.19} & 19.85 & 29.80 & 63.87 & 30.72 & \textbf{83.20} & 50.20 & 41.96 \\
 & \modelnameplsw{} & 22.61 & \textbf{57.53} & \textbf{35.77} & 26.99 & 21.35 & \textbf{31.80} & 63.11 & 30.53 & 81.80 & 51.22 & \textbf{42.27} \\
\midrule
\multirow{3}{*}{3} & PreNorm & 24.40 & 56.23 & 36.08 & 29.58 & 22.43 & 29.80 & \textbf{65.13} & 29.95 & 82.10 & 51.46 & 42.72 \\
 & \modelnamepl{} & \textbf{24.74} & \textbf{58.67} & 36.88 & 29.90 & \textbf{23.04} & \textbf{32.00} & 64.85 & \textbf{30.14} & 84.50 & \textbf{52.17} & \textbf{43.69} \\
 & \modelnameplsw{} & 23.38 & 57.95 & \textbf{37.23} & \textbf{30.70} & 22.24 & 30.60 & 64.85 & 30.05 & \textbf{85.00} & 48.93 & 43.09 \\
\midrule
\multirow{3}{*}{4} & PreNorm & 23.55 & 57.74 & 37.08 & 28.72 & 23.54 & 32.00 & 64.74 & 30.81 & 83.60 & 50.91 & 43.27 \\
 & \modelnamepl{} & 24.32 & \textbf{58.50} & \textbf{38.16} & \textbf{31.21} & 23.75 & \textbf{33.80} & \textbf{64.80} & 30.14 & \textbf{83.90} & \textbf{52.33} & \textbf{44.09} \\
 & \modelnameplsw{} & \textbf{25.26} & 57.83 & 38.12 & 30.31 & \textbf{24.47} & 32.00 & 64.36 & \textbf{31.00} & 82.70 & 51.22 & 43.73 \\
\midrule
\multicolumn{13}{c}{\textbf{370M}} \\
\midrule
\multirow{2}{*}{1} & PreNorm & 26.37 & 62.08 & 43.86 & 35.38 & 28.57 & \textbf{35.40} & \textbf{68.93} & \textbf{32.25} & 85.30 & 51.85 & 47.00 \\
 & \modelnamepl{} & \textbf{28.92} & \textbf{62.33} & \textbf{44.20} & \textbf{36.72} & \textbf{29.90} & 33.20 & 67.63 & 31.48 & \textbf{85.50} & \textbf{54.62} & \textbf{47.45} \\
\midrule
\multirow{3}{*}{2} & PreNorm & \textbf{30.03} & 64.14 & 46.57 & 37.34 & 29.15 & 36.40 & \textbf{70.18} & \textbf{32.63} & \textbf{87.50} & \textbf{55.88} & 48.98 \\
 & \modelnamepl{} & 28.16 & 65.07 & {47.15} & {37.45} & 30.29 & {36.60} & 68.23 & 32.54 & 87.10 & 54.38 & 48.70 \\
 & \modelnameplsw{} & {29.52} & \textbf{65.45} & \textbf{47.75} & \textbf{38.09} & \textbf{32.51} & \textbf{36.60} & 69.31 & 32.44 & {85.80} & 55.01 & \textbf{49.25} \\
\midrule
\multirow{3}{*}{3} & PreNorm & 27.47 & 64.94 & 48.23 & 38.17 & 32.64 & \textbf{38.20} & 68.55 & 31.77 & \textbf{88.90} & 53.99 & 49.29 \\
 & \modelnamepl{} & 29.86 & \textbf{67.42} & 49.63 & 37.76 & \textbf{34.56} & 37.60 & 69.70 & \textbf{32.92} & 88.40 & 54.46 & 50.23 \\
 & \modelnameplsw{} & \textbf{30.89} & 65.82 & \textbf{49.72} & \textbf{40.35} & 34.17 & 37.40 & \textbf{69.80} & \textbf{32.92} & 88.30 & \textbf{57.38} & \textbf{50.67} \\
\midrule
\multirow{3}{*}{4} & PreNorm & 30.46 & 64.90 & 49.31 & 39.12 & 34.74 & 36.80 & 69.10 & 33.88 & 88.00 & \textbf{55.56} & 50.19 \\
 & \modelnamepl{} & \textbf{30.89} & \textbf{65.74} & \textbf{50.92} & 40.29 & \textbf{34.78} & \textbf{39.00} & 69.48 & \textbf{33.97} & \textbf{89.10} & 55.09 & \textbf{50.93} \\
 & \modelnameplsw{} & 30.46 & 65.40 & 50.30 & \textbf{41.80} & 34.68 & 38.40 & \textbf{70.95} & 32.82 & 87.30 & 55.49 & 50.76 \\
\bottomrule
\end{tabular}%
}
\caption{Zero-shot downstream accuracy (\%) for the 140M and 370M models in Table~\ref{tab:ablation}. HellaSwag, OpenBookQA (OBQA), and PIQA use normalized accuracy; all other tasks use accuracy. 
Training and evaluation use the same recurrence $R$; $R=1$ is non-recurrent. \modelnamepl{} and \modelnameplsw{} use residual blocks of 6 sublayers at 140M and 12 at 370M; \modelnameplsw{} retains the input embedding and the three most recent blocks. Bold indicates the highest accuracy within each scale and recurrence group. 
}
\label{tab:controlled_task_accuracy}
\end{table} 

\subsection{Additional analysis of sliding-window depth attention}
\label{app:sliding_window_analysis}
Here we present the complete analysis of sliding-window depth attention, extending Figure~\ref{fig:attnres_140m_r4_sources} to other model scales and recurrence settings.
We also examine whether sliding-window depth attention is similarly effective for standard non-recurrent transformers under the same controlled experimental setup.

\begin{figure}[!ht]
\centering
\includegraphics[width=\linewidth]{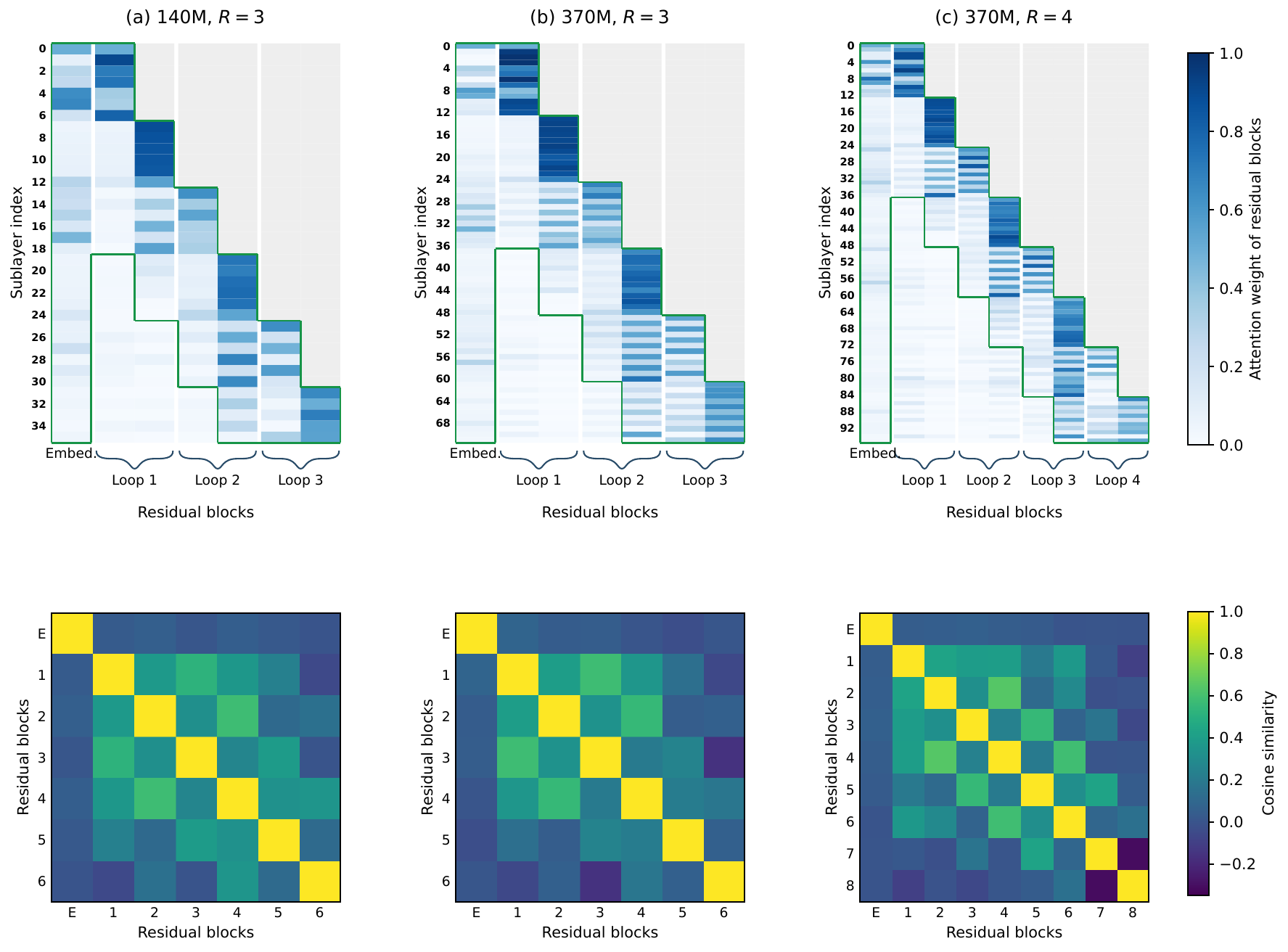}
\caption{Additional full-history diagnostics: (a) 140M, $R=3$; (b) 370M, $R=3$; and (c) 370M, $R=4$. Top: residual-source attention weights, with green outlines marking the embedding plus the three most recent blocks. Bottom: pairwise residual-block cosine similarity, with E denoting the embedding. }
\label{fig:app_routing_cosine}
\label{fig:app_140m_l3}
\label{fig:app_370m_l3}
\label{fig:app_370m_l4}
\end{figure}

\paragraph{Depth-attention weights and residual block similarity across scales.}
Figure~\ref{fig:app_routing_cosine} reports the remaining \modelnamepl{} configurations at 140M, $R=3$, and 370M, $R\in\{3,4\}$. Depth-attention weights are normalized over visible sources for visualization.
Gray cells denote unavailable sources. 
Green outlines identify the sliding window we used in the corresponding \modelnameplsw{}, including the current partial block when present. 
Cosine similarity is computed token-wise between residual-block representations and averaged over 512 validation sequences of length 2,048.
Across model scales and recurrence settings, both the attention weights and pairwise cosine similarities exhibit patterns consistent with those observed in Section~\ref{sec:main_small_results}.

\FloatBarrier

\paragraph{Non-recurrent comparison at matched effective depth.}
\label{app:matched_depth}
We additionally evaluate a non-recurrent transformer with 24 distinct layers, matching the 48 effective sublayers of the six-layer 140M looped model with $R=4$.
It retains the same 768-dimensional hidden states, six attention heads, block size of six sublayers, context length of 2,048, and 11.2B-token training budget.
This comparison therefore matches the effective depth and width of the looped model, but not its parameter count.

\begin{table}[!ht]
\centering
\small
\begin{tabular}{@{}lrrr@{}}
\toprule
Architecture & Full history & Sliding window ($W=3$) & $\Delta$ PPL \\
\midrule
Looped: $L=6$, $R=4$ & 16.07 & 16.03 & $-0.04$ \\
Non-recurrent: $L=24$, $R=1$ & 13.65 & 16.78 & $+3.13$ \\
\bottomrule
\end{tabular}
\caption{Validation perplexity at matched effective depth. Both models use the same training and architecture setups, except that non-recurrence models use more unique parameters to match the effective depth.
The sliding window retains the embedding plus three blocks. 
}
\label{tab:matched_depth_window}
\end{table}

\begin{figure}[!ht]
\centering
\includegraphics[width=.48\linewidth]{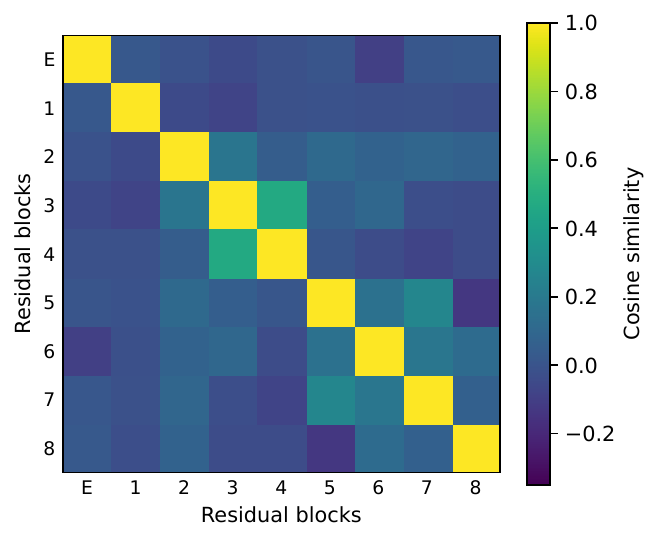}
\caption{Residual-block similarity for the non-recurrent 24-layer, $W=3$ model, measured on the same validation positions as the looped-model cosine diagnostics. E is the embedding; blocks 1--8 belong to one forward pass, not separate recurrences. 
}
\label{fig:app_nonrecurrent_cosine}
\end{figure}

As shown in Table~\ref{tab:matched_depth_window}, sliding-window depth attention slightly improves the looped model, whereas directly applying the same window to the non-recurrent transformer substantially degrades performance.
To better understand this difference, we further visualize the pairwise cosine similarity between residual blocks in the non-recurrent model in Figure~\ref{fig:app_nonrecurrent_cosine}.
The non-recurrent transformer exhibits a markedly different pattern from the looped models, with substantially lower similarity across residual blocks.
This suggests that distinct depths in standard transformers contain less redundant representations, consistent with our hypothesis that the high redundancy across recurrences in looped transformers makes sliding-window depth attention particularly well suited to their recurrent-depth structure.

\subsection{The effect of residual block and sliding window sizes}
\label{app:block_window_sizes}
We additionally study the effect of residual block size, i.e.\ the number of sublayers contained in each residual block, and sliding-window size, i.e.\ the number of recent residual blocks retained by sliding-window depth attention.
We primarily use the 140M \modelname{} for these studies.

Table~\ref{tab:block-size} reports the validation perplexities of models trained with different residual block sizes across recurrence settings.
We find that using finer-grained residual blocks (i.e.\ smaller block sizes) does not consistently improve validation perplexity.
While finer-grained blocks are beneficial for non-recurrent transformers, consistent with \citet{kimi2026attentionresiduals}, we find that their benefit diminishes as the number of recurrences increases, and smaller blocks can even hurt performance at larger $R$.
We hypothesize that this is related to the same phenomenon observed in Section~\ref{sec:main_small_results}: residual-block representations are highly redundant across recurrences, and using smaller blocks introduces more residual sources from earlier recurrences, potentially further increasing this redundancy.

Table~\ref{tab:sliding_window_ppl} reports the results of \modelnameplsw{} with sliding windows retaining the most recent $2$, $3$, $4$, or $5$ residual blocks.
Among the settings we evaluate, retaining three residual blocks performs best overall.
We leave a more systematic exploration of the optimal sliding-window design to future work.

\begin{table}[]
\centering
\setlength{\tabcolsep}{4pt}
\begin{tabular}{llrrr}
\toprule
Model & Recurrence &
\multicolumn{3}{c}{\shortstack{Residual block size\\(sublayers)}} \\
\cmidrule(lr){3-5}
& & 2 & 4 & 6 \\
\midrule
\multirow{4}{*}{\shortstack{140M\\6 layers}}
  & Non-recurrent & \textbf{18.20} & 18.25 & 18.26 \\
  & Loop $=2$     & 17.19 & \textbf{17.18} & 17.27 \\
  & Loop $=3$     & 16.59 & 16.48 & \textbf{16.46} \\
  & Loop $=4$     & 16.13 & \textbf{16.02} & 16.07 \\
\bottomrule
\end{tabular}
\caption{Validation PPL for completed 140M full-history block-size experiments. The best result within each model and recurrence is bolded.}
\label{tab:block-size}
\end{table}

\begin{table}[t]
    \centering
    \small
    \setlength{\tabcolsep}{9pt}
    \begin{tabular}{@{}lcrrrr@{}}
        \toprule
        Model & Recurrence & \multicolumn{4}{c}{Sliding-window blocks $W$} \\
        \cmidrule(lr){3-6}
        & & 2 & 3 & 4 & 5 \\
        \midrule
        140M \modelnameplsw{} & 4 & 16.17 & \textbf{16.03} & 16.07 & 16.14 \\
        \bottomrule
    \end{tabular}
    \caption{Validation PPL versus sliding-window size for 140M \modelnameplsw{} models with residual blocks of six sublayers. $W$ counts retained residual blocks; the input embedding is always retained. 
    }
    \label{tab:sliding_window_ppl}
\end{table}

\FloatBarrier
\section{Large-scale pretraining details}

\subsection{Architecture and Training}
\label{app:data_mixture}
\paragraph{Architecture.}
Table~\ref{tab:large_scale_architecture} details the model backbones used in Section~\ref{sec:scaling_results}.
Compared with the controlled setup, we replace multi-head attention with grouped-query attention to reduce the number of key and value heads, and replace the ReLU$^2$ MLP with SwiGLU.
The full transformer stack is reused at each recurrence.
For both model scales, \modelname{} uses the embedding and the three most recent residual blocks for sliding-window depth attention.
\begin{table}[!ht]
\centering
\small
\setlength{\tabcolsep}{5pt}
\begin{tabular}{@{}lccc@{}}
\toprule
 & \shortstack{PreNorm\\0.6B} & \shortstack{\modelname{}\\0.6B} & \shortstack{\modelname{}\\1.2B} \\
\midrule
Physical Transformer layers $L$ & 18 & 18 & 24 \\
Hidden dimension & 1,536 & 1,536 & 2,048 \\
SwiGLU intermediate dimension & 4,608 & 4,608 & 5,632 \\
Query / key--value heads & 16 / 4 & 16 / 4 & 16 / 4 \\
Head dimension & 96 & 96 & 128 \\
Residual block size (sublayers) & -- & 18 & 24 \\
Fixed recurrence $R$ & 3 & 3 & 3 \\
Effective depth (sublayers) & 108 & 108 & 144 \\
Residual blocks per recurrence & -- & 2 & 2 \\
\midrule
Attention & \multicolumn{3}{c}{Causal grouped-query attention with QK-Norm} \\
MLP & \multicolumn{3}{c}{SwiGLU} \\
Position embeddings & \multicolumn{3}{c}{RoPE ($\theta=50{,}000$)} \\
Tokenizer / vocabulary & \multicolumn{3}{c}{SmolLM2 / 49,152} \\
Normalization & \multicolumn{3}{c}{RMSNorm ($\epsilon=10^{-5}$)} \\
Context length & \multicolumn{3}{c}{4,096} \\
Linear biases & \multicolumn{3}{c}{None} \\
Tied input/output embeddings & \multicolumn{3}{c}{Yes} \\
Depth-attention window & -- & \multicolumn{2}{c}{Embedding + three recent blocks} \\
\midrule
Training tokens & 180B & 180B & 500B \\
Global batch (sequences) & 256 & 256 & 512 \\
Peak / minimum learning rate & $3\!\times\!10^{-3}$ / $3\!\times\!10^{-5}$ & $3\!\times\!10^{-3}$ / $3\!\times\!10^{-5}$ & $2\!\times\!10^{-3}$ / $2\!\times\!10^{-5}$ \\
\bottomrule
\end{tabular}
\caption{Architecture and optimization settings for large-scale pretraining. 
Residual block/window settings apply only to \modelname{}. Each physical layer contains one attention and one MLP sublayer.}
\label{tab:large_scale_architecture}
\end{table}

\paragraph{Pretraining recipe.}
We use AdamW with $(\beta_1,\beta_2)=(0.9,0.95)$, $\epsilon=10^{-8}$, weight decay 0.1, and gradient clipping at 1.0.
A cosine learning-rate schedule follows 1,000 warmup updates and decays to 1\% of the peak learning rate.
Table~\ref{tab:large_scale_architecture} reports the scale-specific peak learning rates and global batch sizes.
Training uses BF16 mixed precision with packed sequences of 4,096 tokens.

\begin{figure}[!ht]
\centering
\includegraphics[width=\linewidth]{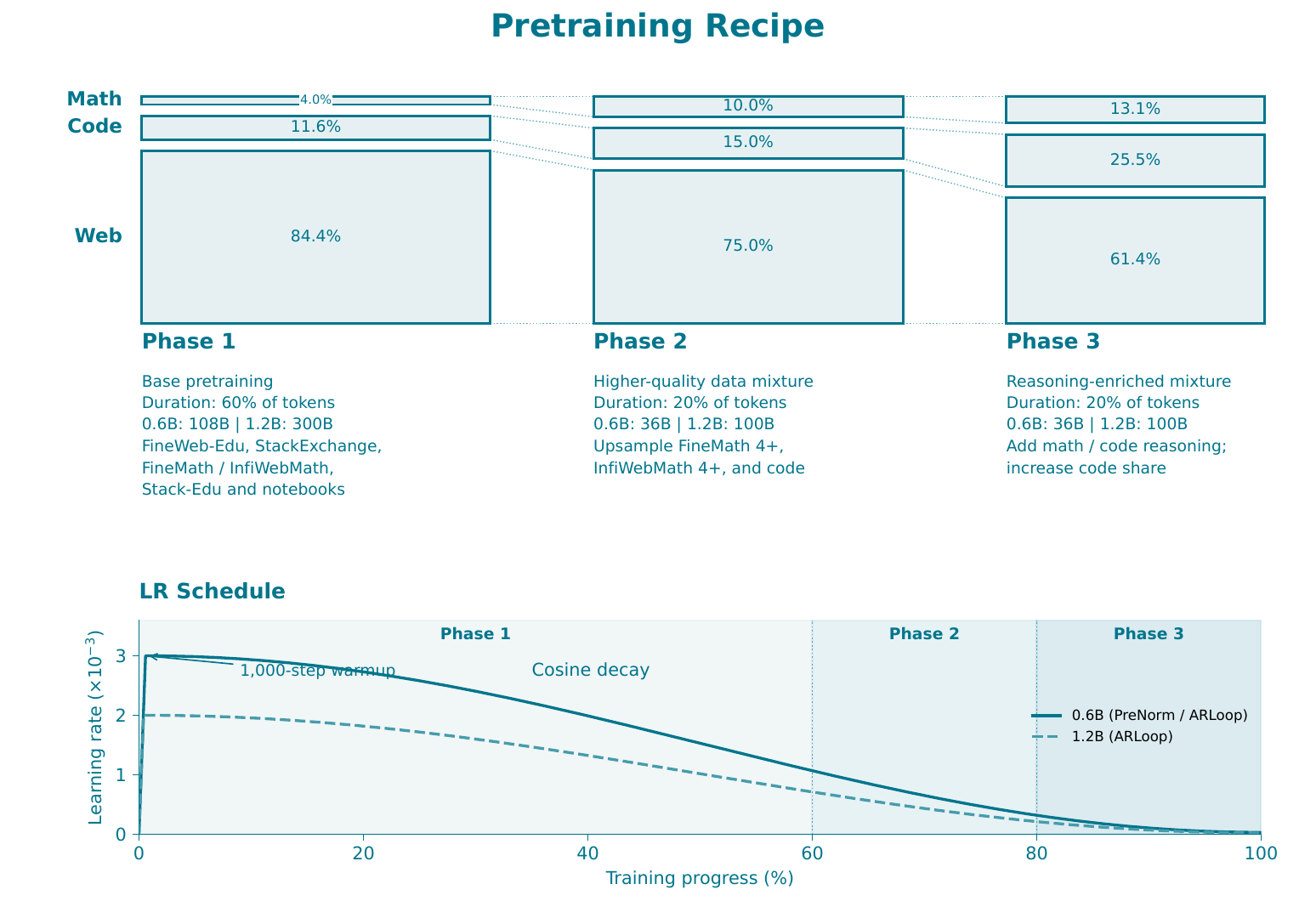}
\caption{Three-stage pretraining recipe. The upper panels show the domain sampling shares at each stage. 
The lower panel shows the learning rate schedules: 1,000 warmup updates followed by cosine decay to 1\% of the peak. 
Source-level allocations are listed in Table~\ref{tab:large_scale_data_mixture}.
}
\label{fig:pretraining_recipe}
\end{figure}

Our pretraining recipe largely follows SmolLM \citep{huggingfacetb2025smollm3} and consists of three training stages, with progressively larger proportions of math and coding data.
Figure~\ref{fig:pretraining_recipe} summarizes how the domain mixture changes throughout training.
The stage boundaries are placed at 60\% and 80\% of the total token budget.
General web data dominate the first stage, while later stages increase the proportions of mathematical and coding data; the final stage additionally introduces reasoning-oriented data sources.
The learning-rate schedule runs continuously across the full training budget, without restarting warmup or cosine decay at stage transitions.

\paragraph{Data source mixture.}
The 0.6B and 1.2B models use the same data-source proportions within each stage, with total training budgets of 180B and 500B tokens, respectively.
Stage~1 accounts for 60\% of each training run, while Stages~2 and 3 each account for 20\%.
Table~\ref{tab:large_scale_data_mixture} reports the resulting per-stage token allocations.
StackExchange is grouped with general data, while notebooks, GitHub issues, OpenCodeReasoning, and Stack-Edu are grouped with coding data.
Dolmino Math aggregates several of its mathematical sub-sources, and Stack-Edu aggregates 15 programming-language subsets.

\subsection{Evaluation setup}
\label{app:large_scale_evaluation}
Table~\ref{tab:large_scale_evaluation_setup} summarizes the benchmark-specific evaluation settings.
For ARC-Challenge, HellaSwag, and WinoGrande, we use few-shot evaluation to follow commonly used evaluation protocols for existing LLMs \citep{zhu2025scaling}.
We use lm-evaluation-harness for evaluating general and mathematical reasoning tasks, and EvalPlus \citep{liu2023codegeneratedchatgptreally} for coding tasks.

For our models, lm-evaluation-harness evaluation is performed in BF16 with a 4,096-token context length, and no chat template.
For mathematical reasoning and coding generation, we use greedy decoding with a maximum of 1,024 new tokens.
GSM8K reports the flexible-extract exact-match metric, while MATH reports Math-Verify scores.
MBPP and HumanEval use the original EvalPlus completion prompts, stop conditions, and sanitization procedures.
For the re-evaluated scores of prior LLMs reported in Table~\ref{tab:scaling_results}, we use each model's native tokenizer and the same task protocols specified in Table~\ref{tab:large_scale_evaluation_setup}.

\begin{table}[!ht]
\centering
\small
\setlength{\tabcolsep}{7pt}
\begin{tabular}{@{}lll@{}}
\toprule
\textbf{Benchmark} & \textbf{Settings / reported metric} & \textbf{Framework} \\
\midrule
\multicolumn{3}{l}{\textbf{General knowledge and reasoning}} \\
ARC-Challenge & Log-likelihood, 25-shot; normalized accuracy & lm-eval-harness \\
HellaSwag & Log-likelihood, 10-shot; normalized accuracy & lm-eval-harness \\
WinoGrande & Log-likelihood, 5-shot; accuracy & lm-eval-harness \\
OpenBookQA & Log-likelihood, 0-shot; normalized accuracy & lm-eval-harness \\
\midrule
\multicolumn{3}{l}{\textbf{Mathematical reasoning}} \\
GSM8K & Generation, 5-shot CoT; flexible-extract EM & lm-eval-harness \\
MATH & Generation, 4-shot Minerva; Math-Verify & lm-eval-harness \\
\midrule
\multicolumn{3}{l}{\textbf{Code generation}} \\
MBPP & Generation, 0-shot; pass@1 & EvalPlus \\
HumanEval & Generation, 0-shot; pass@1 & EvalPlus \\
\bottomrule
\end{tabular}
\caption{Evaluation setup for large-scale pretraining. We evaluate pretrained base checkpoints without chat templates. 
CoT denotes chain-of-thought demonstrations. MATH uses the full 5,000-example test set. 
These settings also apply to the external checkpoints re-evaluated for Table~\ref{tab:scaling_results}.}
\label{tab:large_scale_evaluation_setup}
\end{table}

\begin{table}[t]
    \centering
    \scriptsize
    \setlength{\tabcolsep}{2.2pt}
    \renewcommand{\arraystretch}{0.96}
    \resizebox{\linewidth}{!}{%
    \begin{tabular}{llrrrrrrrr}
        \toprule
        \multirow{2}{*}{Domain} & \multirow{2}{*}{Dataset source}
          & \multicolumn{4}{c}{0.6B model (180B tokens)}
          & \multicolumn{4}{c}{1.2B model (500B tokens)} \\
        \cmidrule(lr){3-6}\cmidrule(lr){7-10}
        & & Stage 1 & Stage 2 & Stage 3 & Total
          & Stage 1 & Stage 2 & Stage 3 & Total \\
        \midrule
        \multirow{2}{*}{General}
          & FineWeb-Edu \citep{penedo2024fineweb}
          & 89.02 & 26.36 & 22.06 & 137.44
          & 247.29 & 73.22 & 61.27 & 381.78 \\
          & StackExchange
          & 2.16 & 0.64 & 0.04 & 2.84
          & 6.00 & 1.78 & 0.10 & 7.88 \\
        \midrule
        \multirow{8}{*}{Math}
          & FineMath 3+ \citep{allal2025smollm2}
          & 1.08 & 0.36 & 0.07 & 1.51
          & 3.00 & 1.00 & 0.20 & 4.20 \\
          & InfiWebMath 3+ \citep{allal2025smollm2}
          & 1.08 & 0.36 & 0.07 & 1.51
          & 3.00 & 1.00 & 0.20 & 4.20 \\
          & FineMath 4+ \citep{allal2025smollm2}
          & 1.08 & 1.96 & 0.90 & 3.94
          & 3.00 & 5.44 & 2.50 & 10.94 \\
          & InfiWebMath 4+ \citep{allal2025smollm2}
          & 1.08 & 0.92 & 0.72 & 2.72
          & 3.00 & 2.56 & 2.00 & 7.56 \\
          & OpenMathReasoning \citep{moshkov2025aimo2}
          & -- & -- & 0.36 & 0.36
          & -- & -- & 1.00 & 1.00 \\
          & Natural Reasoning \citep{yuan2025naturalreasoningreasoningwild28m}
          & -- & -- & 0.36 & 0.36
          & -- & -- & 1.01 & 1.01 \\
          & OpenMathInstruct2 \citep{toshniwal2024openmath2}
          & -- & -- & 2.04 & 2.04
          & -- & -- & 5.66 & 5.66 \\
          & Dolmino Math \citep{olmo2024furious}
          & -- & -- & 0.18 & 0.18
          & -- & -- & 0.51 & 0.51 \\
        \midrule
        \multirow{5}{*}{Coding}
          & OpenCodeReasoning \citep{ahmad2025opencodereasoning}
          & -- & -- & 2.47 & 2.47
          & -- & -- & 6.87 & 6.87 \\
          & Kaggle notebooks \citep{lozhkov2024starcoder}
          & 0.37 & 0.02 & 0.02 & 0.41
          & 1.02 & 0.07 & 0.06 & 1.15 \\
          & Stack v2 Jupyter \citep{lozhkov2024starcoder}
          & 1.08 & 0.36 & 0.54 & 1.98
          & 3.00 & 1.00 & 1.51 & 5.51 \\
          & GitHub Issues \citep{lozhkov2024starcoder}
          & 0.18 & 0.04 & 0.07 & 0.30
          & 0.51 & 0.12 & 0.20 & 0.83 \\
          & Stack-Edu \citep{allal2025smollm2}
          & 10.87 & 4.97 & 6.09 & 21.93
          & 30.19 & 13.82 & 16.91 & 60.91 \\
        \midrule
        \multicolumn{2}{l}{\textbf{Total}}
          & \textbf{108.00} & \textbf{36.00} & \textbf{36.00} & \textbf{180.00}
          & \textbf{300.00} & \textbf{100.00} & \textbf{100.00} & \textbf{500.00} \\
        \bottomrule
    \end{tabular}%
    }
    \caption{Source-level allocation (billion tokens for each cell) for large-scale pretraining. }
    \label{tab:large_scale_data_mixture}
\end{table}

\section{Layer effects and output effects on MBPP and MATH}
\label{app:influence_scores}
We follow the causal-intervention formulation of
\citet{csordas2025languagemodelsusedepth}, adapted to our recurrent architecture.
We compare the final \recurrentSmallModel{} and \prenormSmallModel{}
checkpoints, each with 18 physical layers and three
recurrences, yielding $54$ effective layers.
Within each task, both models are evaluated on the same randomly sampled
50 test examples, drawn from 378 examples for MBPP and 5,000 for MATH.
We use zero-shot prompts and teacher forcing over the canonical solution
code for MBPP or the full reference solution for MATH.
For each unrolled layer $s$, we perform an intervention by ablating its
self-attention and MLP sublayer contributions and rerun the forward pass on the
same input.

\paragraph{Layer effect metric.}
Following \citet{csordas2025languagemodelsusedepth}, we quantify the effect of an
intervention by measuring the change it induces relative to the original run.
Let $\bm{a}_{\ell,t}$ and $\bm{m}_{\ell,t}$ denote the attention and MLP
sublayer outputs, respectively, at unrolled layer $\ell$ and token position
$t$. We define the local layer contribution as
\begin{equation}
    \bm{c}_{\ell,t}
    =
    \bm{a}_{\ell,t} + \bm{m}_{\ell,t}.
\end{equation}
For PreNorm, this corresponds to the additive update contributed by the
transformer layer. We use the same quantity for \modelname{} so that the
two architectures are compared using the same local computation,
independently of how residual sources are subsequently aggregated.

Let $\bar{\bm{c}}_{\ell,t}^{(s)}$ denote the corresponding contribution
after ablating layer $s$. For each later layer $\ell>s$, we measure the
relative change in its computation as
\begin{equation}
    E^{\mathrm{layer}}_{s\rightarrow\ell,t}
    =
    \frac{
        \left\|
        \bm{c}_{\ell,t}
        -
        \bar{\bm{c}}_{\ell,t}^{(s)}
        \right\|_2
    }{
        \left\|\bm{c}_{\ell,t}\right\|_2
    }.
    \label{eq:layer_contribution_influence}
\end{equation}
This measures the effect of intervening on layer $s$ on the local contribution
of a subsequent layer $\ell$.
For an example $e$, let $\mathcal{T}_e$ denote its teacher-forced solution
positions.
For each pair of intervention and later layers $(s,\ell)$, we first take
the maximum $E^{\mathrm{layer}}_{s\rightarrow\ell,t}$ over these positions,
and then the maximum across the 50 examples.
The layer effect of intervention layer $s$ is then obtained by averaging
these values over all strictly later layers.

\paragraph{Output effect metric}
We additionally measure how each intervention affects the model output.
Let $\bm{p}_t$ and $\bar{\bm{p}}_t^{(s)}$ denote the full-vocabulary
probability distributions before and after ablating layer $s$.
Following \citet{csordas2025languagemodelsusedepth}, we measure the output effect
using their $L_2$ distance:
\begin{equation}
    E^{\mathrm{out}}_{s,t}
    =
    \left\|
        \bm{p}_t
        -
        \bar{\bm{p}}_t^{(s)}
    \right\|_2.
\end{equation}
As above, we take the maximum over teacher-forced solution positions and
then across the 50 examples for each intervention layer.

Finally, for both metrics, we normalize the resulting 54-dimensional
effect vector by its sum across intervention layers.
The resulting curves therefore show how the effects are distributed across
the 54 unrolled layers rather than their absolute magnitudes.

\begin{figure}[!ht]
    \centering
    \includegraphics[width=\linewidth]{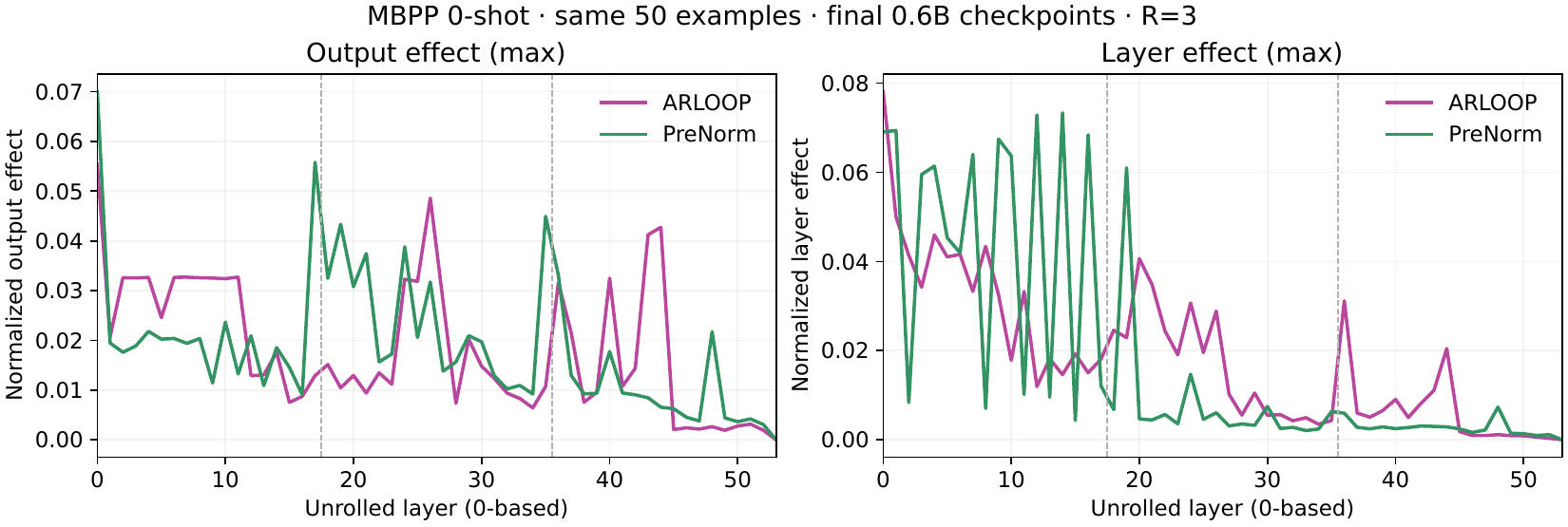}
    \caption{Output influence (left) and layer-contribution influence (right) on 50 shared zero-shot MBPP examples. Curves compare the \recurrentSmallModel{} and \prenormSmallModel{} checkpoints at $R=3$, using max aggregation and normalization by each curve's total influence. Dashed lines separate the three recurrences.}
    \label{fig:mbpp_influence_scores}
\end{figure}

\begin{figure}[!ht]
    \centering
    \includegraphics[width=\linewidth]{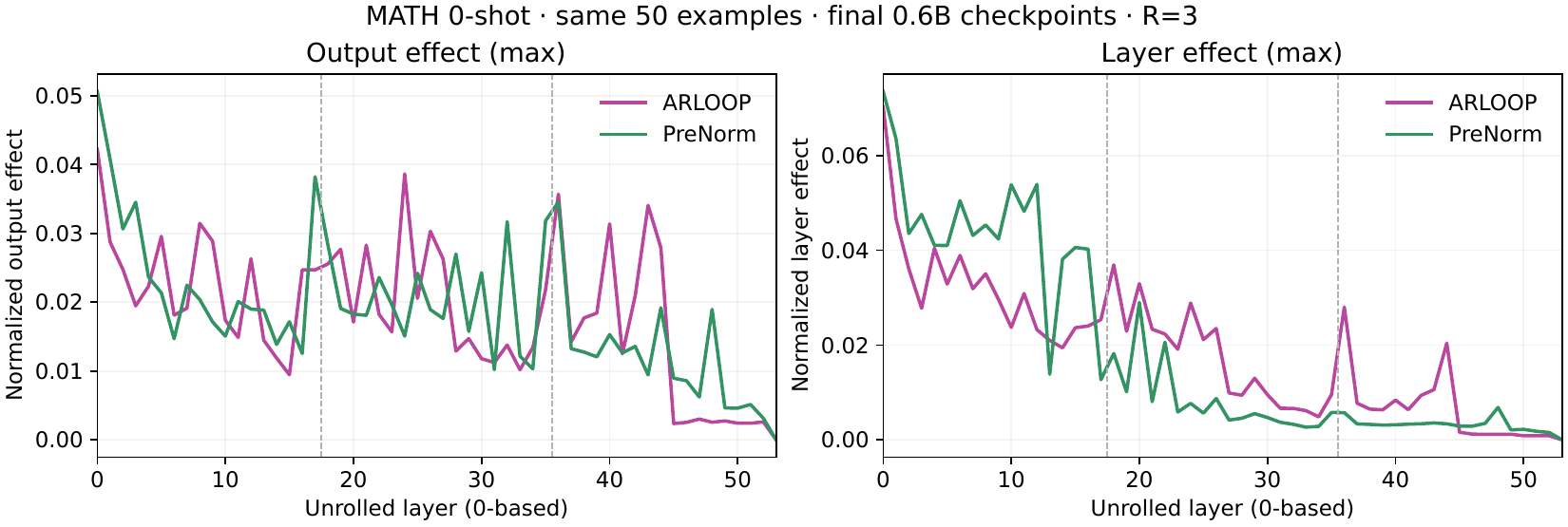}
    \caption{Output influence (left) and layer-contribution influence (right) on 50 shared zero-shot MATH examples, with the same aggregation and normalization as Figure~\ref{fig:mbpp_influence_scores}. The right panel supplies the data for Figure~\ref{fig:math_layer_contribution}. Dashed lines separate the three recurrences.}
    \label{fig:math_influence_scores}
\end{figure}
\FloatBarrier

\section{Dynamic recurrence and recurrence extrapolation}
This section provides the setup and detailed results for the investigations summarized in Section~\ref{sec:discussion}.

\subsection{Training setup and fixed-versus-dynamic results}\label{app:dynamic_recurrence}
\paragraph{Setup.}
The models use the controlled-pretraining architectures and data described in Appendix~\ref{app:exp1_setup}, with 140M and 370M models, except that the recurrence number is not fixed across training.
We consider three recurrence-training strategies:
\modelnamepl{} (i.e.\ fixed recurrences), \poissonSequenceModel{}, and \poissonBatchModel{}.
\modelname{} uses full-history depth attention trained with fixed $R=4$, rather than the default sliding-window variant used for deep recurrence in the main comparison.
For dynamic recurrence, we consider two strategies:
(1) per-sequence recurrence sampled from a Poisson distribution
\citep{prairie2026parcaescalinglawsstable}; and
(2) per-batch recurrence sampled from a Poisson distribution
\citep{geiping2025scaling}.

For the Poisson variants, we sample
\begin{equation}
    \tilde{R} \sim \operatorname{Poisson}(4),
    \qquad
    R = \min\!\left(4,\max\!\left(1,\tilde{R}\right)\right),
    \label{eq:dynamic_poisson}
\end{equation}
In the per-sequence setting, recurrence depth is sampled independently for each sequence,
whereas the per-batch variants use a single sampled depth for the entire batch.
Because per-batch dynamic training requires fewer FLOPs on updates that sample shallow recurrence depths,
we train these models on proportionally more tokens so that the total training FLOPs are matched.
Fixed and per-sequence runs use 11.2B/29.6B tokens at 140M/370M.
Per-batch Poisson runs use 13.6B/36.2B tokens.

\begin{table}[t]
    \centering
    \small
    \setlength{\tabcolsep}{6pt}
    \resizebox{\linewidth}{!}{%
    \begin{tabular}{@{}clrr|rr@{}}
        \toprule
        \multirow{2}{*}{Eval $R$} & \multirow{2}{*}{Training recurrence}
        & \multicolumn{2}{c|}{140M} & \multicolumn{2}{c}{370M} \\
        \cmidrule(lr){3-4}\cmidrule(lr){5-6}
        & & Val PPL $\downarrow$ & downstream avg. $\uparrow$ & Val PPL $\downarrow$ & downstream avg. $\uparrow$ \\
        \midrule
        \multirow{3}{*}{4}
        & \modelnamepl{} ($R=4$) & \textbf{16.07} & \textbf{44.09} & \textbf{12.01} & \textbf{50.93} \\
        & \poissonSequenceModel{} & 16.69 & 43.24 & 12.33 & 49.62 \\
        & \poissonBatchModel{} & 16.52 & 43.30 & 12.36 & 50.53 \\
        \bottomrule
    \end{tabular}%
    }
    \caption{Fixed versus dynamic recurrence training.
    Dynamic Poisson draws have mean 4 and are clamped to $[1,4]$.
    Bold marks the best result at each scale.
    }
    \label{tab:fixed_dynamic_results}
\end{table}

\paragraph{Results.}
Table~\ref{tab:fixed_dynamic_results} presents the results of this comparison.
At evaluation recurrence $R=4$, fixed-depth training achieves the best validation perplexity and downstream average among the tested configurations at each scale.
Hence we adopt fixed recurrences in our main experiment setups.
Despite the negative results of dynamic recurrences, we do not claim that it should be replaced by fixed recurrences.
Such dynamic variants allow flexible computation allocation during inference, and are still a popular choice for looped LLMs.

\subsection{Inference-time recurrence extrapolation}
\label{app:recurrence_extrapolation}
We present the details of the study on inference-time recurrence extrapolation of \modelname{}.

\paragraph{Setup.}
We evaluate the per-sequence Poisson model described above at each recurrence
$R=1,\ldots,8$, using the same validation set as in Section~\ref{sec:exp1}.
During training, the recurrence is bounded by $R\leq4$; evaluation at
$R>4$ therefore requires computation at effective depths that are never
encountered during the original training.
Since our attention-residual mechanism uses a learned query for each
effective sublayer, evaluating beyond the maximum training recurrence
introduces queries at previously unseen depths.
We randomly initialize these additional queries and train only their
parameters on an additional 0.5B FineWeb-Edu tokens, while freezing all
other model parameters.
This provides a favorable setting for recurrence extrapolation: rather
than requiring the model to reuse or extrapolate query parameters to
unseen depths, the newly introduced queries are explicitly optimized to
aggregate the residual history available at those depths.

\begin{figure}[t]
    \centering
    \includegraphics[width=.64\linewidth]{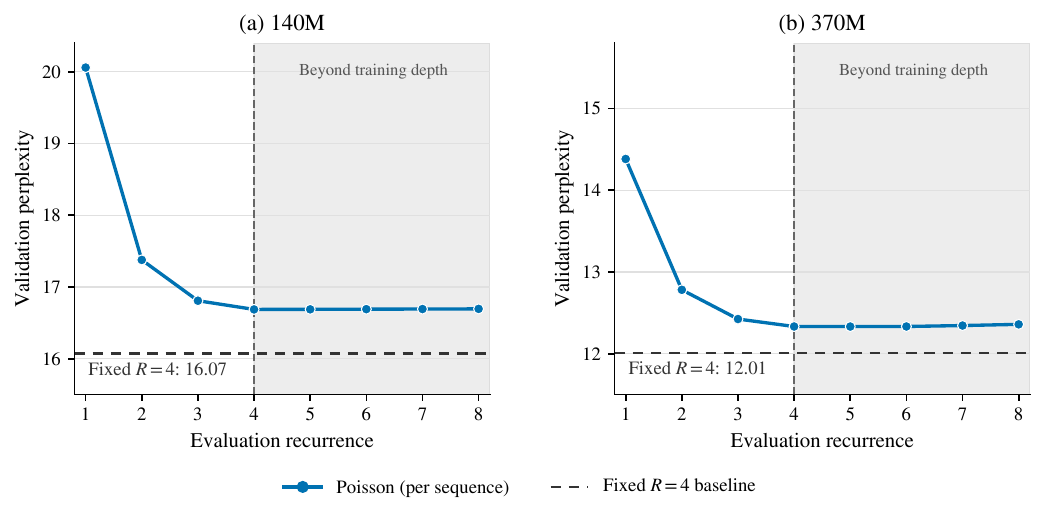}
    \caption{Validation perplexity versus evaluation recurrence for the per-sequence Poisson dynamic full-history \modelname{} models in Table~\ref{tab:fixed_dynamic_results}. Each curve evaluates the same completed checkpoint at $R=1,\ldots,8$. The vertical dashed line at $R=4$ marks the maximum training recurrence; the gray shaded region denotes evaluation beyond this depth. Horizontal dashed lines retain the fixed-$R=4$ baseline reference levels (16.07 at 140M and 12.01 at 370M), not recurrence sweeps of the fixed models. Panels use separate vertical scales.}
    \label{fig:dynamic_recurrence_ppl_full}
\end{figure}

\paragraph{Results.}
Figure~\ref{fig:dynamic_recurrence_ppl_full} shows that perplexity does not
continue to improve beyond the maximum training recurrence at either model
scale.
This remains the case even when the query parameters at unseen depths are
explicitly trained to learn the optimal aggregation of information from preceding depths.
These results support the conclusion in Section~\ref{sec:discussion}:
in our experiments, improved access to residual history alone is insufficient
to produce useful recurrence extrapolation beyond the depths encountered
during training.

\section{Baselines from Prior Work}
\label{app:baselines}

We describe the residual designs used as baselines in
Section~\ref{sec:exp1}, following the notation in Section~\ref{sec:method}.

\paragraph{Sandwich Norm.}
Sandwich Norm \citep{geiping2025scaling,zhu2025scaling}
adds an output normalization to each PreNorm residual update:
\begin{equation}
    h^{(i)} =
    \mathrm{Norm}\left(
        h^{(i-1)} +
        f_i(\mathrm{Norm}(h^{(i-1)}))
    \right)
\end{equation}
This normalizes the residual stream after every sublayer,
preventing uncontrolled growth across recurrent depth.

\paragraph{DeepLoop.}
DeepLoop \citep{li2026deeploopdepthscalinglooped}
extends DeepNorm \citep{wang2022deepnetscalingtransformers1000} to looped transformers by scaling the
skip connection:
\begin{equation}
    h^{(i)} =
    \mathrm{Norm}\left(
        \alpha h^{(i-1)} +
        f_i(\mathrm{Norm}(h^{(i-1)}))
    \right),
\end{equation}
where $\alpha=(2LR)^{1/2}$ for $L$ physical layers
and $R$ recurrences.
DeepLoop further scales the initialization of selected
residual-branch weight matrices by
$\beta=(8LR)^{-1/2}$.
Unlike DeepNorm, which uses an exponent of $1/4$,
DeepLoop uses $1/2$ to account for parameter sharing
across recurrences.

\paragraph{Parcae.}
Parcae \citep{prairie2026parcaescalinglawsstable}
introduces a stable state transition across recurrences.
Let $s^{(r)}=h^{(2Lr)}$ denote the output of recurrence $r$.
The recurrent update takes the form
\begin{equation}
    s^{(r)} =
    \mathcal{S}\left(
        \bar{A}s^{(r-1)} + \bar{B}e
    \right),
\end{equation}
where $e$ denotes the normalized input embedding.
Parcae parameterizes the transition matrices as
\begin{equation}
    A = \mathrm{Diag}(-\exp(a)), \qquad
    \bar{A} = \exp(\Delta A), \qquad
    \bar{B} = \Delta B,
\end{equation}
where $a$, $\Delta$, and $B$ are learnable parameters,
with $\Delta$ constrained to be positive.
This ensures that $\bar{A}$ has eigenvalues smaller
than one in magnitude, stabilizing the recurrent state.
Intuitively, the hidden state is decayed by a learnable
diagonal matrix between recurrences, while the input
embedding is injected into the state before the next recurrence.
The latter mechanism, known as input injection
\citep{bansal2022end}, has been shown to benefit
looped transformers \citep{geiping2025scaling,huang2026loopedmodelsrightii},
but is not used in our \modelname{}.
We evaluate Parcae both with and without input injection
(by setting $\bar{B}=0$ in the latter).
Notably, the original Parcae architecture applies recurrence
only to the middle layers, leaving the initial and final layers
outside the recurrent stack.
For a fair comparison, our implementation instead applies
recurrence to the full transformer stack, consistent with
the other models in our experiments.

\section{Training Cost}
\label{app:training_cost}

We train our models on NVIDIA A100 80GB, H100, and B200 GPUs.
The controlled experiments require approximately 12K GPU hours in total, including early-stage exploration.
Pretraining the \prenormSmallModel{} and \recurrentSmallModel{} models takes approximately one week on 8 H100 GPUs.
The \recurrentLargeModel{} model is trained on 16 B200 GPUs for approximately two weeks.

\end{document}